\documentclass[11pt]{article}

\usepackage[preprint]{acl}

\usepackage{times}
\usepackage{latexsym}

\usepackage{amsthm}
\usepackage[T1]{fontenc}

\usepackage[utf8]{inputenc}

\usepackage{microtype}

\usepackage{inconsolata}

\usepackage{graphicx}

\usepackage{amsmath}   
\usepackage{amssymb}   
\usepackage{amsfonts}

\usepackage{booktabs}
\usepackage{multirow}
\usepackage{xcolor}
\usepackage{colortbl}
\usepackage{tcolorbox}

\usepackage[misc]{ifsym}
\usepackage{fancyhdr}

\makeatletter
\fancypagestyle{firstpage}{
  \fancyhf{}
  \renewcommand{\headrulewidth}{0.4pt}
  \IfFileExists{hunyuanlogo.png}{
    \lhead{\includegraphics[height=18pt]{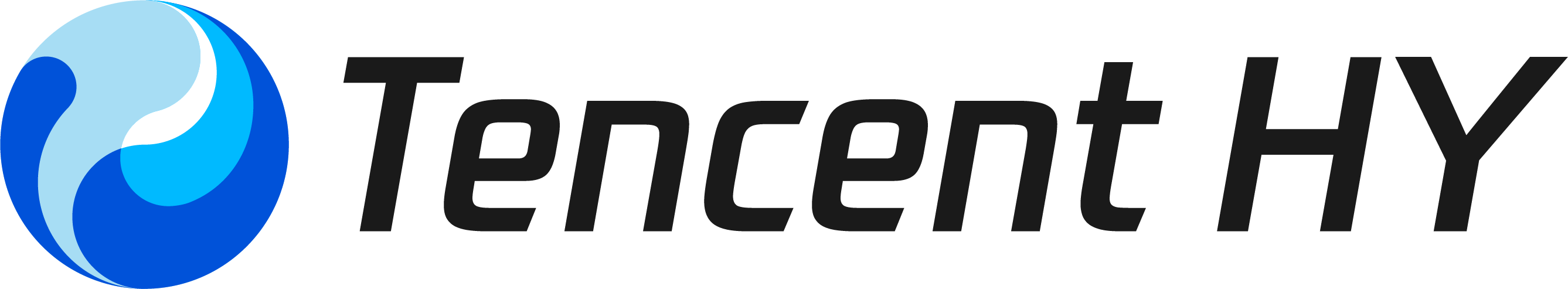}}
  }{}
  \rhead{\textcolor[HTML]{4D4D4D}{\small \today}}
  \renewcommand{\headrule}{{\color[HTML]{4D4D4D}\hrule\@height\headrulewidth\@width\headwidth\vskip-\headrulewidth}}
  \fancyfoot[C]{\thepage}
}
\makeatother

\title{ORBIT: On-policy Exploration-Exploitation \\ for Controllable Multi-Budget Reasoning}

\author{
    \textbf{Kun Liang}$^{1,2,*}$ \quad
    \textbf{Clive Bai}$^{3}$ \quad
    \textbf{Xin Xu}$^{3,4}$ \quad
    \textbf{Chenming Tang}$^{1,2,*}$ \\
    \textbf{Sanwoo Lee}$^{1,2,*}$ \hspace{0.4em}
    \textbf{Weijie Liu}$^{3}$ \hspace{0.4em}
    \textbf{Saiyong Yang}$^{3,\dagger}$ \hspace{0.4em}
    \textbf{Yunfang Wu}$^{1,2,\dagger}$ \\
    \noalign{\vskip 5pt}
    $^1$School of Computer Science, Peking University \\
    $^2$National Key Laboratory for Multimedia Information Processing, Peking University \\
    $^3$LLM Department, Tencent \\
    $^4$The Hong Kong University of Science and Technology \\
    \noalign{\vskip 5pt}
    \Letter~{\fontsize{10.5}{12}\selectfont\texttt{kliang25@stu.pku.edu.cn}} \hspace{0.5em}
    \Letter~{\fontsize{10.5}{12}\selectfont\texttt{wuyf@pku.edu.cn}}
}

\begin{document}
\maketitle

\thispagestyle{firstpage}
\pagestyle{plain}

\renewcommand*{\thefootnote}{\fnsymbol{footnote}}
\footnotetext{$^*$ Work done during internship at Tencent.}
\footnotetext{$^\dagger$ Corresponding Authors.}
\renewcommand*{\thefootnote}{\arabic{footnote}}

\begin{abstract}
Recent Large Reasoning Models (LRMs) achieve strong performance by leveraging long-form Chain-of-Thought (CoT) reasoning, but uniformly applying overlong reasoning at inference time incurs substantial and often unnecessary computational cost. 
To address this, prior work explores various strategies to infer an appropriate reasoning budget from the input.
However, such approaches are unreliable in the worst case, as estimating the minimal required reasoning effort is fundamentally difficult, and the trade-off between reasoning cost and accuracy is implicitly fixed during training, limiting flexibility under varying deployment scenarios.
Motivated by these limitations, we propose \textbf{ORBIT}, a controllable multi-budget reasoning framework with well-separated reasoning modes triggered by prompt.
ORBIT employs multi-stage reinforcement learning to discover Pareto frontier reasoning behaviors at each effort, followed by On-Policy Distillation to fuse these behaviors into a single unified model. 
Experiments show that ORBIT achieves (1) controllable reasoning behavior over multiple modes, (2) competitive reasoning density within each mode, and (3) integration of these frontier policies into a single unified student model while preserving clear mode separation and token-efficient per-mode performance.

\end{abstract}

\section{Introduction}
\label{sec:intro}

\begin{figure*}[t]
    \centering
    \includegraphics[width=1.05\linewidth]{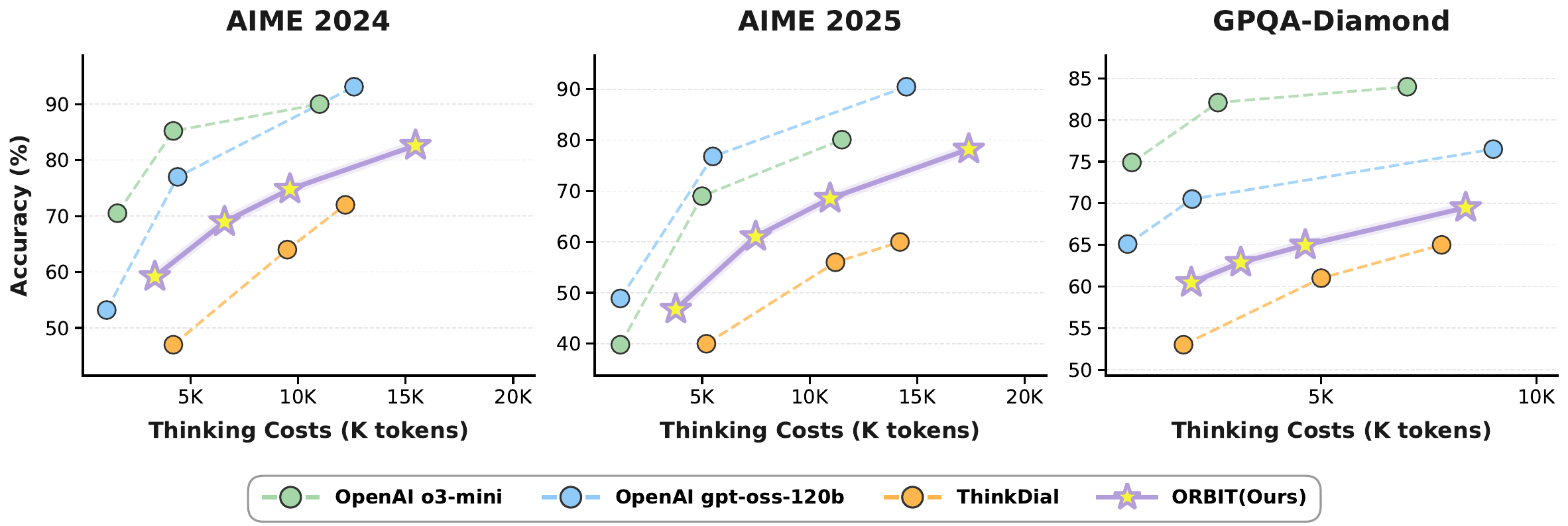}

    \caption{ORBIT demonstrates controllable reasoning behaviors comparable to advanced systems, covering multiple reasoning modes: low/mid/high for o3-mini and gpt-oss-120B, and an extended set of low/mid/high/extra-high for ORBIT, illustrating its ability to flexibly trade off reasoning cost and performance across modes.}

    \label{fig:orbit_against_sota}
\end{figure*}

Recent advances in Reinforcement Learning with Verifiable Rewards (RLVR)~\citep{guo2025deepseekr1,team2025kimi,xiaomi2025mimo} have significantly pushed the frontier of complex task solving and long-form Chain-of-Thought (CoT)~\citep{weichainofthought} reasoning in large language models. By encouraging explicit reasoning trajectories, these methods enable models to explore richer solution paths and achieve strong performance on challenging benchmarks~\citep{guo2025deepseekr1}. However, the continued push toward greater reasoning depth introduces a practical trade-off between reasoning effort and operational efficiency: increasingly long reasoning traces incur higher latency and computational cost, while often being unnecessary for simpler queries~\citep{yang2025thinking}.

Notably, this observation concerns not the use of long-form reasoning during training, but its uncalibrated application at inference time, as longer reasoning chains often facilitate exploration of richer solution paths and contribute to higher peak accuracy~\citep{setlur2025e3}. The challenge primarily emerges at deployment time, where models trained to reason deeply may apply the same heavy reasoning patterns indiscriminately across inputs.

% 这里校对一下相关工作
To optimize inference-time efficiency, prior work has explored various strategies for adjusting reasoning effort according to task demands~\citep{luo2025adar1hybridcotbileveladaptive,lou2025adacot,luo2025o1prunerlengthharmonizingfinetuningo1like,zhang2025adaptthinkreasoningmodelslearn}. Across these approaches, the underlying goal is essentially the same: to estimate an optimal allocation of reasoning effort. 

However, this estimation problem is fundamentally difficult. Determining the minimal amount of reasoning required to solve an instance can be viewed as attempting to estimate its Kolmogorov complexity~\citep{10.5555/1478784}, which is known to be uncomputable and connected to the halting problem theoretically~\citep{graves2017adaptivecomputationtimerecurrent}.
Empirically, adaptive mechanisms can also misallocate reasoning effort—truncating reasoning on instances that would benefit from deeper deliberation~\citep{aggarwal2025optimalthinkingbenchevaluatingunderthinkingllms,jin2025wellthinkingenhancingllm}, while preserving unnecessarily long reasoning for simpler instances~\citep{chen2025think23overthinkingo1like}.

Moreover, these inference-agnostic approaches implicitly determine the trade-off between reasoning effort and expected accuracy during training, reflecting a fixed preference over cost and performance. While suitable for standardized benchmarks, this design may be less flexible in deployment scenarios where user preferences and system constraints vary across queries.

Motivated by these limitations, we rethink a controllable reasoning model that can flexibly adjust its reasoning effort according to user preferences by switching reasoning policies across modes at inference time, while achieving competitive reasoning density within each effort level.

We introduce \textbf{ORBIT}, a fully on-policy training framework for controllable multi-budget reasoning. By exploring under different training context windows, 
ORBIT identifies a Pareto frontier of reasoning behaviors and effectively integrates these into a single model whose discrete reasoning modes can be triggered by specific prompts. As illustrated in Figure~\ref{fig:orbit_against_sota}, ORBIT achieves clear and well-separated reasoning modes similar to state-of-the-art controllable systems, while maintaining strong per-mode accuracy and token-efficient reasoning.

Our contributions are as follows: 
\begin{enumerate}
    \item We propose \textbf{ORBIT}, a fully on-policy training framework for controllable multi-budget reasoning, enabling a single model to generate responses across multiple reasoning modes triggered by mode-specific prompts.
    \item We introduce an \textbf{Expansion-Compression Loop} exploration strategy that efficiently discovers mode-specific reasoning frontiers under varying context constraints, yielding high quality policies at each reasoning effort.
    \item We design a \textbf{Multi-Teacher On-Policy Distillation} pipeline with mode-aware initialization, which fuses these frontier policies into a single unified student model while preserving clear mode separation and token-efficient performance across all modes.
\end{enumerate}

\section{Preliminary}
\label{sec:pre}

\paragraph{Problem Statement}

We study controllable multi-budget reasoning, aiming to learn a single, unified policy 
\(\pi_\phi(o \mid q, c)\) that generates distinct, mode-specific reasoning behaviors for a query 
\(q \in \mathcal{D}\), where the mode \(c \in \mathcal{C} = \{c_1, \dots, c_K\}\) can be selected 
or triggered at inference time, corresponding to different levels of reasoning effort.

Crucially, each induced behavior is expected to be \emph{strong in its own right}:
for every mode $c_k$, the conditional policy
$\pi_\phi(\cdot \mid q, c_k)$
should achieve high task accuracy relative to its reasoning budget, rather than representing a compromised linear interpolation performance.

The core challenge, therefore, lies in learning such a unified yet controllable policy under a fixed training budget, while maintaining clear behavioral separation across modes.

\paragraph{RLVR}

We model language generation as autoregressive sampling from a conditional probability distribution $\pi_{\theta}(o| q)$, and our training goal is to maximize the expected correctness reward $r(q,o) \in \{0,1\}$: 
\begin{equation}
\max_{\theta} \mathbb{E}_{q \sim \mathcal{D},\, o \sim \pi_{\theta}} [r(q, o)],
\end{equation}
where $q$ represents the input query and $o$ represents the output response.

In this work, we adopt Group Relative Policy Optimization (GRPO) \citep{shao2024deepseekmath} as our RLVR algorithm for its simplicity and effectiveness in training reasoning models \citep{guo2025deepseekr1}.
Unlike traditional methods \citep{schulman2017proximal}, GRPO bypasses the value model by computing advantages relative to a group of $G$ responses $\{o_1, \dots, o_G\}$ sampled from $\pi_{\theta_\text{old}}$. The objective is:

\begin{multline}
\mathcal{J}_{\text{GRPO}}(\theta)
= \mathbb{E}_{q \sim \mathcal{D}}
\Biggl[
\frac{1}{G} \sum_{i=1}^G \frac{1}{|o_i|} 
\sum_{t=1}^{|o_i|} \\
\min \Bigl(
    r_{i,t}(\theta) \hat{A}_{i,t},
    \text{clip}(r_{i,t}(\theta), 1-\epsilon, 1+\epsilon)\hat{A}_{i,t}
\Bigr)
\Biggr],
\end{multline}

where the importance ratio $r_{i,t}(\theta)$ and the group-relative advantage $\hat{A}_i$ are defined as:

{
\begin{align}
&r_{i,t}(\theta) = \frac{\pi_\theta(o_{i,t}|q, o_{i,<t})}{\pi_{\theta_\text{old}}(o_{i,t}|q, o_{i,<t})},\\&\hat{A}_i = \frac{r(q, o_i) - \text{mean}(\{r(q, o_j)\})}{\text{std}(\{r(q, o_j)\})}.    
\end{align}}

\section{Methodology}
\label{sec:method}

\begin{figure*}[t]
\centering
\includegraphics[width=1\linewidth]{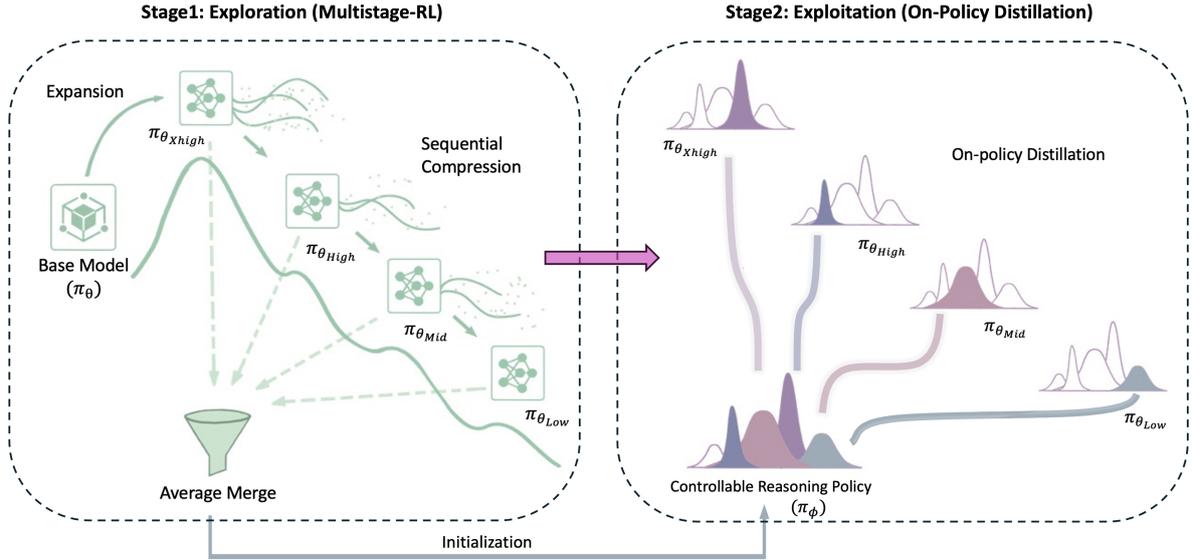}
\caption{The ORBIT framework overview: (1) In the \textbf{Exploration} stage, multi-stage RL discovers specialized reasoning frontiers under varying context from the Expansion-Compression Loop. (2) In the \textbf{Exploitation} stage, these behaviors are unified into a single student model via Model Merging and Multi-Teacher On-Policy Distillation.}
\label{fig:overview}
\end{figure*}

We approach controllable multi-budget reasoning through an \emph{exploration--exploitation} perspective. 
To operationalize, we decompose the problem into two complementary phases:

\textbf{Exploration Phase.}
Multi-stage RL allocates computational effort to sample sequences in semantic space under varying context budgets, discovering high-quality per-mode strategies.

\textbf{Exploitation Phase.}
Model merging and On-Policy Distillation consolidate these policies into a single student model, preserving mode-specific behaviors while enabling an efficient unified model for all reasoning modes.

Figure~\ref{fig:overview} provides an overview of this framework.

\subsection{Expansion-Compression Loop for Mode-Specific Policy Exploration}
\label{sec:exploration}

During exploration, each reasoning mode \(c \in \mathcal{C}\) is subject to a context budget reflecting its intended reasoning effort. The training objective is to discover high-quality strategies that maximize task accuracy under a given budget. To achieve this, we adopt the simple length regularizing method--truncation, along with a scheduling strategy and formalize the exploration stage as traversal along the accuracy--length Pareto frontier.

\paragraph{Dual-Objectivity under Truncation}

For a given context budget $L_k$ and reasoning mode $c \in \mathcal{C}$, we consider a policy $\pi_{\theta_k}$ and define its truncated, mode-specific version:

\begin{equation}
\tilde{\pi}_{\theta_k}^{(c)}(o; L_k) := \pi_{\theta_k}(o \mid q,c)\, \mathbb{I}(|o| \le L_k),
\end{equation}

For a given mode $c$, the expected reward (accuracy) under truncation is
\begin{equation}
J^{(c)}(\pi_{\theta_k})
=
\mathbb{E}_{q \sim \mathcal{D}}
\mathbb{E}_{o \sim \tilde{\pi}_{\theta_k}^{(c)}(\cdot \mid q)}
[r(q,o)],
\end{equation}

We define the non-truncation probability
\begin{equation}
p_{\text{non-trunc}}^{\pi,(c)}(q)
:= 
\mathbb{P}_{o \sim \pi_{\theta_k}(\cdot \mid q, c)}(|o| \le L_k),
\end{equation}
and the corresponding renormalized truncated policy
\begin{equation}
\pi_{\theta_k}^{T_k,(c)}(o \mid q) := 
\frac{\pi_{\theta_k}(o \mid q, c)\, \mathbb{I}(|o| \le L_k)}
     {p_{\text{non-trunc}}^{\pi,(c)}(q)}.
\end{equation}

Then the truncated expected reward decomposes as
\begin{align}
J^{(c)}(\pi_{\theta_k}) &= 
\mathbb{E}_{q \sim \mathcal{D}}
\Big[
p_{\text{non-trunc}}^{\pi,(c)}(q)\,
\bar{r}_{\text{non-trunc}}^{\pi,(c)}(q)
\Big],\\
\bar{r}_{\text{non-trunc}}^{\pi,(c)}(q) &:= 
\mathbb{E}_{o \sim \pi_{\theta_k}^{T_k,(c)}(\cdot \mid q)} [r(q,o)].
\end{align}

This decomposition shows that, for each mode $c$, optimization under truncation involves two coupled directions:
\begin{itemize}
    \item increasing $p_{\text{non-trunc}}^{\pi,(c)}$, i.e., concentrating probability mass on valid-length trajectories;
    \item increasing $\bar{r}_{\text{non-trunc}}^{\pi,(c)}$, i.e., improving the semantic quality of non-truncated trajectories.
\end{itemize}

\paragraph{Harvesting Along the Frontier via an Expansion--Compression Loop}

To efficiently obtain per-mode strategies under truncation, we introduce a one-time \textbf{Expansion--Compression Loop} for scheduling context window in RLVR training:

\begin{itemize}
    \item \textbf{Expansion (Semantic Optimality).} 
We first enlarge the context budget to $L_{\mathrm{exp}} \gg \mathbb{E}[|o^+|]$, where $o^+$ denotes correct, within-window trajectories, ensuring truncation is inactive and the policy focuses on semantic correctness. This allows exploration of harder instances and establishes a global performance ceiling.

\item \textbf{Sequential Compression (Density Optimality).}

Starting from the expanded policy, we progressively tighten the context budget $L_k$. This constraint imposes truncation, inducing a discontinuous length distribution that can destabilize training through two mechanisms: (i) increased reward variance due to the scarcity of valid-length correct trajectories, which leads to noisy gradient estimates; and (ii) ``survival mode,'' where the model collapses to the shortest, trivial solutions, suppressing the complex reasoning required for harder instances.

To balance mode separability and training stability, we adopt a logarithmic compression schedule ($L_{k+1}=L_k/2$) and sequentially optimize the policy under progressively smaller budgets. This schedule applies comparable truncation pressure on the remaining trajectory mass at each stage, allowing the policy to re-equilibrate under the current constraint before the next compression. Further discussion of this design is provided in Appendix \ref{app:discussion}.

Notably, binary compression serves as the canonical instantiation of our framework. If finer-grained trade-offs between output quality and latency are desired, additional compression stages can be simply inserted into the current schedule.

\end{itemize}

The loop yields stage-wise per-mode experts along the accuracy--length frontier, which provide policy anchors in semantic space for consolidating multiple reasoning modes into a single model.

\subsection{On-Policy Distillation for Frontier Policy Exploitation}
\label{sec:exploitation}

While the exploration stage identifies frontier policies, REINFORCE-style updates provide only a sparse $O(1)$ supervision signal per rollout sequence (uniform $\hat{A}$ across all tokens).
Such scalar feedback is insufficient not only for efficient consolidation, but also for reasoning behaviors anchoring across multiple mode-specific experts.

Transitioning from \textit{discovery} to \textit{consolidation}, we therefore adopt distillation as a high-bandwidth alternative, enabling supervision at a tokenwise $O(N)$ resolution. 
Specifically, we treat the frontier expert models obtained during the exploration stage, $\{\pi_{\theta_{k}}\}_{k=1}^K$, as teachers and fuse their mode-specific behaviors into a single unified student model $\pi_\phi$. This is achieved through a unified pipeline that combines \textbf{Multi-Teacher On-Policy Distillation} with a \textbf{Mode-aware Initialization}, allowing diverse reasoning modes to be consolidated in a structured and controllable manner.

\paragraph{On-Policy Distillation Mode Fusion}

A common intuition is to perform distillation using an offline dataset of query–response pairs sampled from the teacher model, and then train the student in an SFT manner.

However, distilling an autoregressive student model in this way leads to a training--inference distribution mismatch, commonly known as \emph{exposure bias} \citep{bengio2015scheduled}, because supervision is provided on teacher-generated trajectories that the student policy may never enter during inference, preventing it from learning to \emph{calibrate its own behavior} on the trajectories induced by its own policy~\citep{agarwal2023gkd}.

Based on these intuitions, we adopt \textbf{On-Policy Distillation}~\citep{agarwal2023gkd,lu2025onpolicydistillation}, minimizing the Reverse Kullback-Leibler (RKL) divergence:

\begin{equation}
    \mathcal{L}_{\text{distill}}(\phi) = \mathbb{E}_{q \sim \mathcal{D}, {o} \sim \pi_\phi(\cdot|q, p_k)} \left[ \log \frac{\pi_\phi({o}|q, p_k)}{\pi_{\theta_k}({o}|q, p_k)} \right].
\end{equation}

RKL is inherently \textbf{mode-seeking}, encouraging the student $\pi_\phi$ to focus on the high-density reasoning modes of the expert $\pi_{\theta_{k}}$ under prompt $p_k$ for mode $c_k$. 
By concentrating probability mass on these modes, the student is less likely to assign weight to irrelevant trajectories or non-target ones. This ensures that the reasoning knowledge from each expert is effectively consolidated, while preserving the model’s broader generative capabilities.

Formally, the objective function is defined as
\begin{multline}
\label{equ:okd}
    \mathcal{L}_{\text{fusion}}(\phi) = \mathbb{E}_{k \sim U(1, K)} \mathbb{E}_{\substack{q \sim \mathcal{D} \\ {o} \sim \pi_\phi(\cdot|q, p_k)}} \\
    \left[ \sum_{t} \log \frac{\pi_\phi(o_t | q, p_k, o_{<t})}{\pi_{\theta_k}(o_t | q, p_k,  o_{<t})} \right]
\end{multline}

\paragraph{Mode-aware initialization via Model Merging}  
Despite the theoretical advantages of RKL, On-Policy Distillation suffers a "cold-start": if the student $\pi_\phi$ initially lacks confidence in the target modes, rollouts rarely reach the expert’s high-density manifold, limiting the supervisory signal~\citep{jang2026stable}. We mitigate this by initializing the student via model merging, combining the stage-wise frontier models $\{\pi_{\theta_k}\}_{k=1}^K$ obtained during exploration, to align input-output mappings with the teacher $\pi_{\theta_{k}}$. This provides a more stable start and accelerates convergence. 
% under our RKL distillation setting
Details can be found in Appendix~\ref{app:model_merge}.

\section{Experiments}
\subsection{Experimental Settings}

We implemented ORBIT on three models— \texttt{DeepSeek-\allowbreak Distill-\allowbreak Qwen-\allowbreak 1.5B}, 
\texttt{Qwen3-\allowbreak 4B-\allowbreak 2507-\allowbreak Thinking}, and \texttt{Openmath-\allowbreak Nemotron-\allowbreak 7B}—each configured with four reasoning modes: Low, Mid, High, and Xhigh. The training context lengths for these modes were set according to the model: 2K, 4K, 8K, and 16K tokens for \texttt{DeepSeek-\allowbreak Distill-\allowbreak Qwen-\allowbreak 1.5B}, and 4K, 8K, 16K, and 32K tokens for both \texttt{Qwen3-\allowbreak 4B-\allowbreak 2507-\allowbreak Thinking} and \texttt{Openmath-\allowbreak Nemotron-\allowbreak 7B}. Specifically, \texttt{DeepSeek-Distill-Qwen-1.5B} was trained on the DAPO-Math-17K dataset~\citep{yu2025dapo}, while \texttt{Qwen3-\allowbreak 4B-\allowbreak 2507-\allowbreak Thinking} and \texttt{Openmath-\allowbreak Nemotron-\allowbreak 7B} were trained on Polaris-53K~\citep{Polaris2025}. Notably, the training procedure is scheduled in reverse order. Further implementation and training details are provided in Appendix~\ref{app:train_detail}.

\textbf{Baselines.}  
We first evaluate our full pipeline against the recent controllable reasoning models, including o3-mini~\citep{o3mini}, gpt-oss-120b~\citep{gpt-oss}, and ThinkDial~\citep{heThinkDialOpenRecipe2025}. Next, we compare ORBIT's within-mode performance against current RL-based adaptive and efficient reasoning methods~\citep{aggarwal2025l1controllinglongreasoning,arora2025traininglanguagemodelsreason,zhang2025adaptthinkreasoningmodelslearn,fang2025thinklessllmlearnsthink,tu2025learningthinkshapingadaptive,liu2025learnreasonefficientlyadaptive} on \texttt{DeepSeek-\allowbreak Distill-\allowbreak Qwen-\allowbreak 1.5B}. Finally, we ablate ORBIT with other potentially used methods and baselines.

\textbf{Evaluation Benchmarks.}  
We evaluate our methods on three mathematical reasoning tasks: AIME 2024 \& 2025, BeyondAIME~\citep{bytedance_seed_2025_beyondaime}, as well as representative multi-task benchmarks spanning diverse domains in science, humanities, and professional knowledge: GPQA-Diamond~\citep{gpqa} and MMLU-Pro~\citep{mmlu-pro}.

\newcommand{\method}{ORBIT}

\definecolor{lightblue1}{RGB}{240, 248, 255}
\definecolor{lightblue2}{RGB}{220, 240, 255}
\definecolor{lightblue3}{RGB}{200, 230, 255}
\definecolor{lightblue4}{RGB}{180, 220, 255}
\definecolor{lightblue5}{RGB}{160, 210, 255}

\definecolor{lightgreen1}{RGB}{240, 255, 240}
\definecolor{lightgreen2}{RGB}{220, 245, 220}
\definecolor{lightgreen3}{RGB}{200, 235, 200}
\definecolor{lightgreen4}{RGB}{180, 225, 180}
\definecolor{lightgreen5}{RGB}{160, 215, 160}

\definecolor{lightcyan1}{RGB}{240, 255, 255}
\definecolor{lightcyan2}{RGB}{220, 250, 250}
\definecolor{lightcyan3}{RGB}{200, 245, 245}
\definecolor{lightcyan4}{RGB}{180, 240, 240}
\definecolor{lightcyan5}{RGB}{160, 235, 235}

\definecolor{lightpurple1}{RGB}{245, 240, 255}
\definecolor{lightpurple2}{RGB}{235, 220, 255}
\definecolor{lightpurple3}{RGB}{225, 200, 255}
\definecolor{lightpurple4}{RGB}{215, 180, 255}
\definecolor{lightpurple5}{RGB}{205, 160, 255}
\definecolor{lightpurple6}{RGB}{195, 140, 255}

\definecolor{lightred1}{RGB}{255, 240, 240}
\definecolor{lightred2}{RGB}{255, 220, 220}
\definecolor{lightred3}{RGB}{255, 200, 200}
\definecolor{lightred4}{RGB}{255, 180, 180}
\definecolor{lightred5}{RGB}{255, 160, 160}
\definecolor{lightred6}{RGB}{255, 140, 140}

\definecolor{lightorange1}{RGB}{255, 245, 230}
\definecolor{lightorange2}{RGB}{255, 235, 210}
\definecolor{lightorange3}{RGB}{255, 225, 190}
\definecolor{lightorange4}{RGB}{255, 215, 170}
\definecolor{lightorange5}{RGB}{255, 205, 150}
\definecolor{lightorange6}{RGB}{255, 195, 130}

\definecolor{lightyellow1}{RGB}{255, 255, 240}
\definecolor{lightyellow2}{RGB}{255, 250, 210}

\definecolor{lightyellow3}{RGB}{255, 245, 180}
\definecolor{lightyellow4}{RGB}{255, 240, 150}

\begin{table*}[!t]
    \centering
    % removed duplicate label to avoid conflicts
    % \vspace{2mm}
    \resizebox{0.95\textwidth}{!}{
    \begin{tabular}{l|cc|cc|cc|cc|cc|cc}
    		    \toprule[1.6pt]
        & \multicolumn{6}{c|}{\textbf{Mathematics}} & \multicolumn{4}{c|}{\textbf{Multi-Task}} & \multicolumn{2}{c}{\multirow{2}{*}{\textbf{Overall}}} \\
        \cmidrule(lr){2-11}
        \multirow{3}{*}{\textbf{Models}} 
           & \multicolumn{2}{c|}{\textbf{AIME 24}} 
           & \multicolumn{2}{c|}{\textbf{AIME 25}} 
           & \multicolumn{2}{c|}{\textbf{Beyond AIME}} 
           & \multicolumn{2}{c|}{\textbf{GPQA}} 
           & \multicolumn{2}{c|}{\textbf{MMLU}} 
           & \multicolumn{2}{c}{} \\ \cmidrule(lr){2-13}
        & \texttt{avg@32} & \texttt{Toks}
        & \texttt{avg@32} & \texttt{Toks}
        & \texttt{avg@8} & \texttt{Toks}
        & \texttt{avg@8} & \texttt{Toks}
        & \texttt{avg@1} & \texttt{Toks}
        & \texttt{Avg.} & \texttt{Toks} \\
        \midrule
        \multicolumn{13}{c}{\texttt{\textbf{{DeepSeek-Distill-Qwen-1.5B}}}} \\
        \midrule
        Initial Model & 28.6 & 18.8 & 22.8 & 18.5 & 10.1 & 18.6 & 28.0 & 12.7 & 35.5 & 7.2 & 25.0 & 15.2 \\
        - {\method} Low & \cellcolor{lightpurple1}{28.7} & \cellcolor{lightorange4}{2.7} & \cellcolor{lightpurple1}{21.1} & \cellcolor{lightorange4}{2.3} & \cellcolor{lightpurple1}{9.9} & \cellcolor{lightorange4}{1.9} & \cellcolor{lightpurple2}{34.8} & \cellcolor{lightorange4}{1.8} & \cellcolor{lightpurple1}{38.1} & \cellcolor{lightorange4}{1.4} & \cellcolor{lightpurple1}{26.5} & \cellcolor{lightorange4}{2.0} \\
        - {\method} Mid & \cellcolor{lightpurple2}{34.5} & \cellcolor{lightorange3}{5.6} & \cellcolor{lightpurple2}{26.0} & \cellcolor{lightorange3}{4.9} & \cellcolor{lightpurple2}{12.6} & \cellcolor{lightorange3}{3.6} & \cellcolor{lightpurple1}{34.3} & \cellcolor{lightorange3}{2.6} & \cellcolor{lightpurple2}{38.2} & \cellcolor{lightorange3}{2.0} & \cellcolor{lightpurple2}{29.1} & \cellcolor{lightorange3}{3.7} \\
        - {\method} High & \cellcolor{lightpurple3}{38.8} & \cellcolor{lightorange2}{7.7} & \cellcolor{lightpurple3}{30.2} & \cellcolor{lightorange2}{7.4} & \cellcolor{lightpurple3}{15.6} & \cellcolor{lightorange2}{7.1} & \cellcolor{lightpurple3}{41.0} & \cellcolor{lightorange2}{4.9} & \cellcolor{lightpurple4}{41.7} & \cellcolor{lightorange2}{3.4} & \cellcolor{lightpurple3}{33.5} & \cellcolor{lightorange2}{6.1} \\
        - {\method} Xhigh & \cellcolor{lightpurple4}{42.6} & \cellcolor{lightorange1}{8.5} & \cellcolor{lightpurple4}{31.1} & \cellcolor{lightorange1}{7.8} & \cellcolor{lightpurple4}{15.9} & \cellcolor{lightorange1}{7.7} & \cellcolor{lightpurple4}{41.8} & \cellcolor{lightorange1}{6.4} & \cellcolor{lightpurple3}{41.2} & \cellcolor{lightorange1}{4.4} & \cellcolor{lightpurple4}{34.5} & \cellcolor{lightorange1}{7.0} \\
        \midrule
        \multicolumn{13}{c}{\texttt{\textbf{{Qwen3-4B-2507-Thinking}}}} \\
        \midrule
        Initial Model & 81.6 & 20.1 & 77.1 & 18.3 & 52.4 & 27.6 & 66.1 & 8.8 & 74.4 & 4.4 & 70.3 & 15.8 \\
        - {\method} Low & \cellcolor{lightpurple1}{59.7} & \cellcolor{lightorange4}{3.3} & \cellcolor{lightpurple1}{46.8} & \cellcolor{lightorange4}{3.8} & \cellcolor{lightpurple1}{31.4} & \cellcolor{lightorange4}{3.4} & \cellcolor{lightpurple1}{60.5} & \cellcolor{lightorange4}{2.0} & \cellcolor{lightpurple1}{71.7} & \cellcolor{lightorange4}{1.1} & \cellcolor{lightpurple1}{54.0} & \cellcolor{lightorange4}{2.7} \\
        - {\method} Mid & \cellcolor{lightpurple2}{69.5} & \cellcolor{lightorange3}{6.6} & \cellcolor{lightpurple2}{61.0} & \cellcolor{lightorange3}{7.5} & \cellcolor{lightpurple2}{40.6} & \cellcolor{lightorange3}{7.4} & \cellcolor{lightpurple2}{62.9} & \cellcolor{lightorange3}{3.1} & \cellcolor{lightpurple2}{72.8} & \cellcolor{lightorange3}{1.9} & \cellcolor{lightpurple2}{61.4} & \cellcolor{lightorange3}{5.3} \\
        - {\method} High & \cellcolor{lightpurple3}{75.0} & \cellcolor{lightorange2}{9.6} & \cellcolor{lightpurple3}{69.2} & \cellcolor{lightorange2}{11.0} & \cellcolor{lightpurple3}{48.3} & \cellcolor{lightorange2}{11.8} & \cellcolor{lightpurple3}{65.2} & \cellcolor{lightorange2}{4.6} & \cellcolor{lightpurple3}{73.5} & \cellcolor{lightorange2}{2.8} & \cellcolor{lightpurple3}{66.2} & \cellcolor{lightorange2}{8.0} \\
        - {\method} Xhigh & \cellcolor{lightpurple4}{82.6} & \cellcolor{lightorange1}{15.5} & \cellcolor{lightpurple4}{78.2} & \cellcolor{lightorange1}{17.4} & \cellcolor{lightpurple4}{53.5} & \cellcolor{lightorange1}{20.7} & \cellcolor{lightpurple4}{69.5} & \cellcolor{lightorange1}{8.4} & \cellcolor{lightpurple4}{74.5} & \cellcolor{lightorange1}{5.4} & \cellcolor{lightpurple4}{71.7} & \cellcolor{lightorange1}{13.5} \\
        \midrule
        \multicolumn{13}{c}{\texttt{\textbf{{Openmath-Nemotron-7B}}}} \\
        \midrule
        Initial Model & 71.6 & 11.9 & 59.9 & 13.7 & 41.8 & 13.6 & 32.5 & 8.5 & 42.4 & 5.7 & 49.6 & 10.7 \\
        - {\method} Low & \cellcolor{lightpurple1}{60.0} & \cellcolor{lightorange4}{5.2} & \cellcolor{lightpurple1}{36.5} & \cellcolor{lightorange4}{5.7} & \cellcolor{lightpurple1}{31.5} & \cellcolor{lightorange4}{4.7} & \cellcolor{lightpurple1}{31.5} & \cellcolor{lightorange4}{2.4} & \cellcolor{lightpurple1}{39.3} & \cellcolor{lightorange4}{1.8} & \cellcolor{lightpurple1}{39.8} & \cellcolor{lightorange4}{4.0} \\
        - {\method} Mid & \cellcolor{lightpurple2}{66.4} & \cellcolor{lightorange3}{7.4} & \cellcolor{lightpurple2}{45.2} & \cellcolor{lightorange3}{8.3} & \cellcolor{lightpurple2}{35.0} & \cellcolor{lightorange3}{7.5} & \cellcolor{lightpurple2}{32.3} & \cellcolor{lightorange3}{4.2} & \cellcolor{lightpurple2}{41.5} & \cellcolor{lightorange3}{2.9} & \cellcolor{lightpurple2}{44.1} & \cellcolor{lightorange3}{6.1} \\
        - {\method} High & \cellcolor{lightpurple3}{71.4} & \cellcolor{lightorange2}{10.0} & \cellcolor{lightpurple3}{57.1} & \cellcolor{lightorange2}{11.6} & \cellcolor{lightpurple3}{41.4} & \cellcolor{lightorange2}{10.9} & \cellcolor{lightpurple1}{31.9} & \cellcolor{lightorange2}{6.1} & \cellcolor{lightpurple3}{41.7} & \cellcolor{lightorange2}{4.4} & \cellcolor{lightpurple3}{48.7} & \cellcolor{lightorange2}{8.6} \\
        - {\method} Xhigh & \cellcolor{lightpurple4}{72.0} & \cellcolor{lightorange1}{11.0} & \cellcolor{lightpurple4}{57.8} & \cellcolor{lightorange1}{12.1} & \cellcolor{lightpurple4}{41.8} & \cellcolor{lightorange1}{12.1} & \cellcolor{lightpurple4}{34.8} & \cellcolor{lightorange1}{6.6} & \cellcolor{lightpurple4}{42.2} & \cellcolor{lightorange1}{5.0} & \cellcolor{lightpurple4}{49.7} & \cellcolor{lightorange1}{9.4} \\
        \bottomrule[1.6pt]
    \end{tabular}}
    \caption{Comparison of reasoning modes of {\method} across various reasoning tasks.
    \label{tab:orbit_overview}
``\texttt{Avg@k}'' denotes the average accuracy (in \%) over k generations (i.e., \texttt{pass@1}), and "\texttt{Toks}" indicates the average output length in thousands of tokens (K).
Darker colors in the cell background denote better results.
    }
%\vspace{-0.5em}
\end{table*}

\subsection{Main Results}
\paragraph{ORBIT Achieves Controllable Reasoning Behavior Across Domains}

Table~\ref{tab:orbit_overview} illustrates the reasoning behaviors learned by ORBIT under different reasoning budgets across domains. We observe a clear and consistent separation between reasoning modes, with each exhibiting a monotonic trade-off between reasoning effort and task performance. This separation indicates that ORBIT successfully learns prompt triggered inference-time separable distinct reasoning policies.

Importantly, these controllable behaviors transfer reliably across domains. When transferring from mathematical reasoning to general question answering benchmarks, the ordinal ordering of reasoning modes is preserved, while only minimal stepwise performance degradation is incurred.

\paragraph{ORBIT Achieves Competitive Token Efficiency Within Each Mode}

\definecolor{lightred1}{RGB}{255, 240, 240}
\definecolor{lightred2}{RGB}{255, 220, 220}
\definecolor{lightred3}{RGB}{255, 200, 200}
\definecolor{lightred4}{RGB}{255, 180, 180}
\definecolor{lightred5}{RGB}{255, 160, 160}
\definecolor{lightred6}{RGB}{255, 140, 140}

\definecolor{lightpurple1}{RGB}{245, 240, 255}
\definecolor{lightpurple2}{RGB}{235, 220, 255}
\definecolor{lightpurple3}{RGB}{225, 200, 255}
\definecolor{lightpurple4}{RGB}{215, 180, 255}
\definecolor{lightpurple5}{RGB}{205, 160, 255}
\definecolor{lightpurple6}{RGB}{195, 140, 255}

\definecolor{lightorange1}{RGB}{255, 245, 230}
\definecolor{lightorange2}{RGB}{255, 235, 210}
\definecolor{lightorange3}{RGB}{255, 225, 190}
\definecolor{lightorange4}{RGB}{255, 215, 170}
\definecolor{lightorange5}{RGB}{255, 205, 150}
\definecolor{lightorange6}{RGB}{255, 195, 130}

\begin{table*}[!t]
    \centering
    % \vspace{2mm}
    \resizebox{0.95\textwidth}{!}{
    \begin{tabular}{l|cc|cc|cc|cc|cc|cc}
        \toprule[1.6pt]
        \multirow{3}{*}{\textbf{Models}} 
           & \multicolumn{2}{c|}{\textbf{AIME 24}} 
           & \multicolumn{2}{c|}{\textbf{AIME 25}} 
           & \multicolumn{2}{c|}{\textbf{Beyond AIME}} 
           & \multicolumn{2}{c|}{\textbf{GPQA}} 
           & \multicolumn{2}{c|}{\textbf{MMLU}} 
           & \multicolumn{2}{c}{\textbf{Overall}} \\ \cmidrule(lr){2-13}
        & \texttt{avg@32} & \texttt{Toks}
        & \texttt{avg@32} & \texttt{Toks}
        & \texttt{avg@8} & \texttt{Toks}
        & \texttt{avg@8} & \texttt{Toks}
        & \texttt{avg@1} & \texttt{Toks}
        & \texttt{Avg.} & \texttt{Toks} \\ 
        \midrule
        Initial Model & 28.6 & 18.8 & 22.8 & 18.5 & 10.1 & 18.6 & 28.0 & 12.7 & 35.5 & 7.2 & 25.0 & 15.2 \\ 
        \hline
        1) TLMRE-DS-1.5B ($\alpha=0.1$) & 28.1 & 14.7 & 22.1 & 14.2 & 8.4 & 12.7 & 30.8 & 7.8 & 34.3 & 4.7 & 24.7 & 10.8 \\
        2) AdaptThink-1.5B ($\delta=0.05$) & 25.2 & \cellcolor{lightorange3}{7.6} & 21.3 & 8.1 & 9.9 & \cellcolor{lightorange3}{5.1} & 31.3 & 6.7 & 36.9 & 4.1 & 24.9 & 6.3 \\
        3) AutoThink-DS-1.5B-Stage3 & \cellcolor{lightpurple1}{31.0} & 11.0 & \cellcolor{lightpurple1}{22.8} & 10.6 & \cellcolor{lightpurple2}{11.0} & 9.4 & 31.4 & 7.6 & 36.3 & 4.5 & 26.5 & 8.6 \\
        4) Laser-DE-L4096-1.5B & 30.6 & 8.2 & 22.0 & \cellcolor{lightorange2}{7.5} & 10.3 & 7.2 & \cellcolor{lightpurple1}{34.6} & \cellcolor{lightorange3}{3.4} & 37.8 & \cellcolor{lightorange3}{2.3} & \cellcolor{lightpurple1}{27.1} & \cellcolor{lightorange3}{5.7} \\
        \hline
        5) AutoThink-Stage3$^*$ & \cellcolor{lightpurple4}{39.5} & 9.0 & \cellcolor{lightpurple3}{29.0} & 9.0 & \cellcolor{lightpurple2}{12.4} & 8.0 & \cellcolor{lightpurple3}{35.7} & \cellcolor{lightorange2}{4.7} & \cellcolor{lightpurple3}{39.6} & 3.6 & \cellcolor{lightpurple3}{31.2} & 6.9 \\
        6) L1-1.5B-Max$^*$ & 23.6 & \cellcolor{lightorange5}{3.4} & 22.3 & \cellcolor{lightorange5}{3.1} & 10.4 & \cellcolor{lightorange5}{3.1} & 33.0 & \cellcolor{lightorange6}{1.2} & \cellcolor{lightpurple2}{39.5} & \cellcolor{lightorange6}{1.4} & 25.8 & \cellcolor{lightorange5}{2.4} \\
        7) Thinkless-1.5B-RL$^*$ & 27.8 & 11.1 & 21.8 & 11.1 & \cellcolor{lightpurple2}{11.1} & 11.6 & 28.9 & 12.8 & 35.2 & 6.6 & 25.0 & 10.6 \\
        \hline
        - ORBIT Low & \cellcolor{lightpurple1}{28.7} & \cellcolor{lightorange6}{2.7} & \cellcolor{lightpurple1}{21.1} & \cellcolor{lightorange6}{2.3} & \cellcolor{lightpurple1}{9.9} & \cellcolor{lightorange6}{1.9} & \cellcolor{lightpurple2}{34.8} & \cellcolor{lightorange5}{1.8} & \cellcolor{lightpurple1}{38.1} & \cellcolor{lightorange6}{1.4} & \cellcolor{lightpurple1}{26.5} & \cellcolor{lightorange6}{2.0} \\
        - ORBIT Mid & \cellcolor{lightpurple2}{34.5} & \cellcolor{lightorange4}{5.6} & \cellcolor{lightpurple2}{26.0} & \cellcolor{lightorange4}{4.9} & \cellcolor{lightpurple3}{12.6} & \cellcolor{lightorange4}{3.6} & \cellcolor{lightpurple1}{34.3} & \cellcolor{lightorange4}{2.6} & \cellcolor{lightpurple2}{38.2} & \cellcolor{lightorange4}{2.0} & \cellcolor{lightpurple2}{29.1} & \cellcolor{lightorange4}{3.7} \\
        - ORBIT High & \cellcolor{lightpurple3}{38.8} & \cellcolor{lightorange2}{7.7} & \cellcolor{lightpurple4}{30.2} & \cellcolor{lightorange3}{7.4} & \cellcolor{lightpurple4}{15.6} & \cellcolor{lightorange2}{7.1} & \cellcolor{lightpurple4}{41.0} & 4.9 & \cellcolor{lightpurple4}{41.7} & \cellcolor{lightorange2}{3.4} & \cellcolor{lightpurple4}{33.5} & \cellcolor{lightorange2}{6.1} \\
        - ORBIT Xhigh & \cellcolor{lightpurple4}{42.6} & 8.5 & \cellcolor{lightpurple4}{31.1} & 7.8 & \cellcolor{lightpurple4}{15.9} & 7.7 & \cellcolor{lightpurple4}{41.8} & 6.4 & \cellcolor{lightpurple3}{41.2} & 4.4 & \cellcolor{lightpurple4}{34.5} & 7.0 \\
        \bottomrule[1.6pt]
    \end{tabular}}
    \caption{Comparison of reasoning modes of ORBIT trained from \texttt{DeepSeek-\allowbreak Distill-\allowbreak Qwen-\allowbreak 1.5B} with efficient reasoning baselines across various reasoning tasks. Each reasoning mode is evaluated and compared individually.  
% Darker colors in the cell background denote better results.
    }
    \label{tab:ds_adaptive}
\end{table*}

We evaluate each ORBIT mode separately on \texttt{DeepSeek-\allowbreak Distill-\allowbreak Qwen-\allowbreak 1.5B}, comparing its accuracy–length trade-off with RL-based efficient reasoning methods. We aim to determine if the multi-mode student maintains high reasoning density across all triggered efforts without performance compromise.

As shown in Table~\ref{tab:ds_adaptive}, ORBIT consistently establishes a superior accuracy–length Pareto frontier,  reaching higher accuracy under the same reasoning budget or matching baseline accuracy with fewer tokens. Notably, these results are obtained with a minimalist truncation-based reward combined with our scheduling and fusion strategy, highlighting the effectiveness and simplicity of ORBIT.

\section{Analysis}

We conducted a series of analytical experiments on the best-performing \texttt{Qwen3-\allowbreak 4B-\allowbreak 2507-\allowbreak Thinking}.

\begin{figure*}
    \centering
    \includegraphics[width=\linewidth]{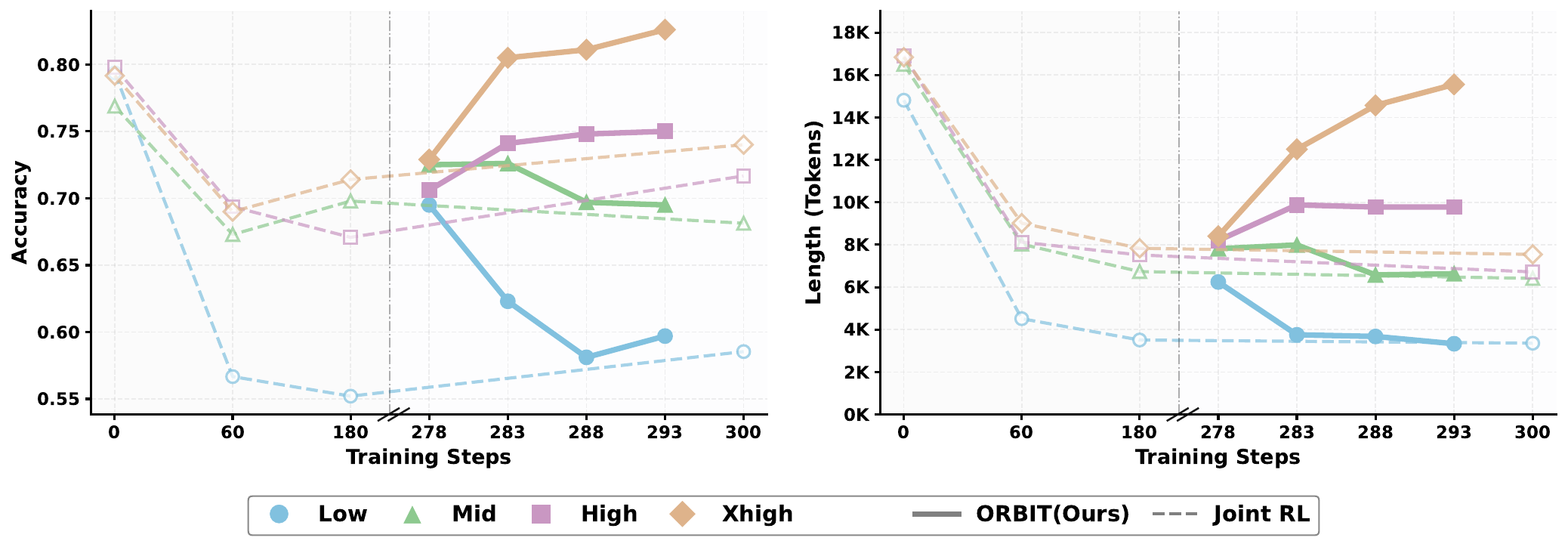}
    \caption{Step-wise performance of ORBIT and joint RL on AIME24, with steps aligned by total training tokens for fair comparison. ORBIT starts from step 278, where OPD training begins. Sequential compression stages are omitted for clarity.}
    \label{fig:ablation_parallel_rl}
\end{figure*}

\paragraph{Comparison with End-to-End Joint RL Training}  
We first investigate whether mode separation can be integrated into the RL training, where we jointly train each single query under four different reasoning modes, each associated with the specific system prompt and corresponding context window. To ensure the peak performance, we start our joint RL training after expansion stage, and stabilize the training with rejection sampling to prevent training oscillation under the most aggressive truncation (e.g. 32K $\rightarrow$ 4K). All other training settings are identical to those of our sequential counterpart.

As shown in Figure~\ref{fig:ablation_parallel_rl}, joint RL leads to mode collapse, where the Mid, High, and Xhigh modes converge to nearly identical behaviors.

We identify two key factors contributing to this: 
(i) supervision from truncated sequences dominates optimization, which is consistent with our truncation-rate decomposition in Sec.~\ref{sec:exploration}; and 
(ii) higher-mode sequences can satisfy their truncation requirements by compressing below lower modes. 
Together, these effects encourage reward hacking on response length rather than the intended mode-selective reasoning behavior.

\begin{figure*}
    \centering
    \includegraphics[width=\linewidth]{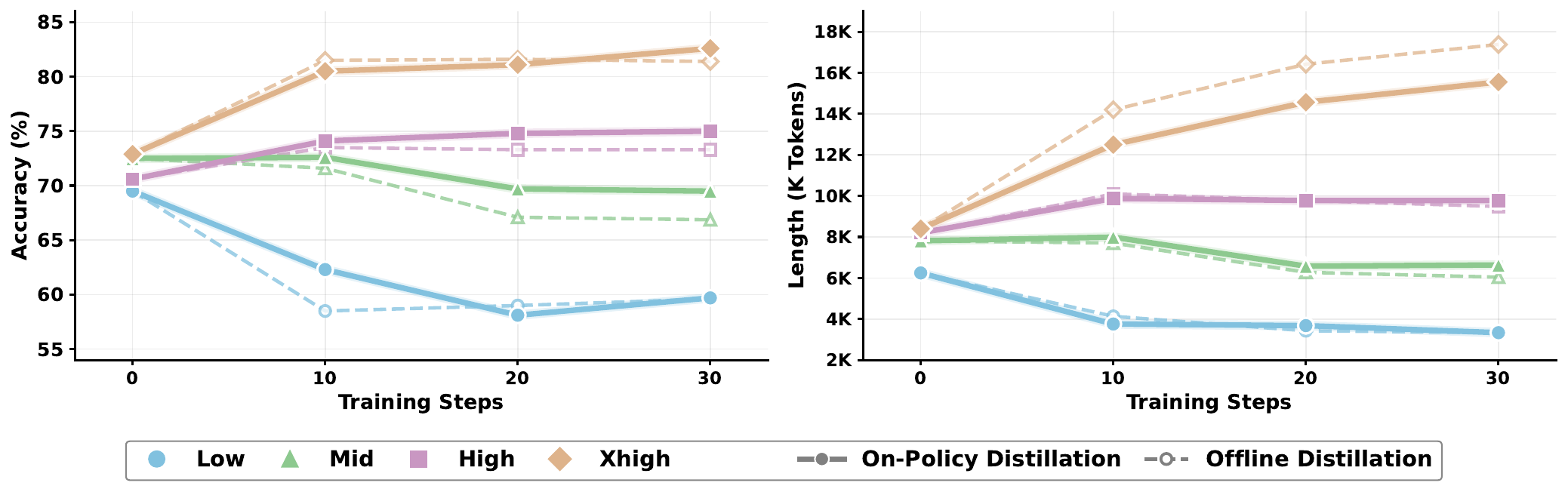}
    \caption{Comparison of OPD and offline distillation as the fusing strategy on AIME24, with steps aligned by total training samples for fair comparison. On-policy fusion achieves slightly higher performance while both strategies show similar trends.}
    \label{fig:ablation_offline_sft}
\end{figure*}

\paragraph{Analysis of Fusion Strategies}  
We examine ORBIT's fusion strategy by replacing on-policy fusion with offline distillation, where consolidation proceeds without additional on-policy rollouts, relying instead on a pre-sampled dataset from the exploration-stage frontier models. Both variants are initialized from the same merged model to minimize behavioral inconsistency.

As shown in Figure~\ref{fig:ablation_offline_sft}, on-policy fusion achieves slightly higher performance than offline distillation, while both methods exhibit similar convergence trends. This outcome can be attributed to the strong framework-level alignment in ORBIT: all teacher and student models share a sequential training with same base model trained on same data with same algorithm, effectively reducing policy mismatch and rendering offline distillation in this setting closer to a semi-offline procedure.

Beyond performance, on-policy fusion provides operational advantages: it generates rollouts dynamically, adapts convergence on-the-fly, and can naturally integrate with RL-based correctness signals. In contrast, offline distillation decouples dataset construction from training, allowing for scalable evaluation via pre-collected datasets, which can be leveraged in techniques like Rejection Fine-tuning. 
The choice between these strategies, or their combination, can be tailored depending on specific application requirements.

\paragraph{ORBIT's Soft-Mode Constraint Elicits Cognition Separation}

\begin{figure}
    \centering
    \includegraphics[width=0.95\linewidth]{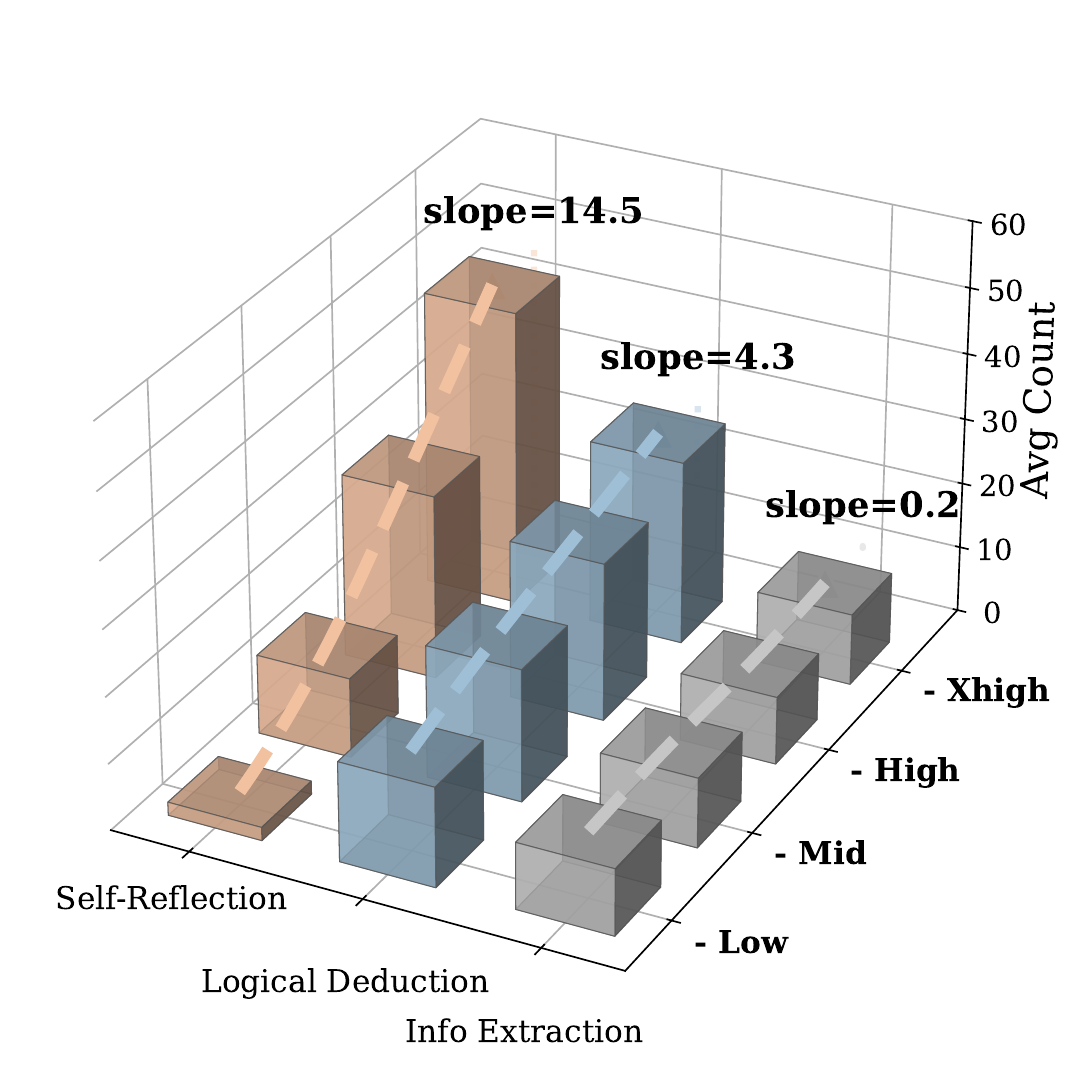}
    \caption{Cognition analysis of ORBIT across reasoning modes, vertical axis denotes average count per sample.}
    \label{fig:cognitive}
\end{figure}

To further analyze how ORBIT's mode control shapes reasoning behavior, we evaluate it on MATH500~\citep{math500} and visualize the evolution of reasoning components in Figure~\ref{fig:cognitive}. 

We categorize ORBIT’s reasoning behavior along three functional levels:
(i) \emph{Logical Deduction} steps grow linearly as modes increase, reflecting structured problem-solving calibrated by training context;
(ii) \emph{Self-Reflection} exhibits characteristics analogous to System 2 thinking via an exponential-like increase, allocating computational effort toward iterative reasoning as modes advance; 
and (iii) \emph{Information Extraction} remains constant, suggesting that increased token usage stems from deeper reasoning rather than additional input parsing. 

Together, these trends suggest a structured and hierarchical organization of reasoning behavior, in which different reasoning components occupy distinct functional levels and higher reasoning modes systematically amplify the contributions of more reflective, high-level processes.

\section{Related Work}

\paragraph{Reasoning Effort Adjustment in LRMs}

Existing works on reasoning effort adjustment primarily target overthinking through RL-based length penalties~\citep{aggarwal2025l1controllinglongreasoning,arora2025traininglanguagemodelsreason,yi2025shorterbetter,zhang2025adaptthinkreasoningmodelslearn, fang2025thinklessllmlearnsthink, lou2025adacot}, verification training~\citep{chen2025verithinkerlearningverifymakes}, early exit strategies~\citep{yang2025dynamicearlyexitreasoning,jiang2025flashthinkearlyexitmethod}, or inference-time interventions~\citep{wang2025waitdontneedwait}. Few lines of methods tackle the underthinking problem by decoding time intervention~\citep{muennighoff2025s1simpletesttimescaling,jin2025wellthinkingenhancingllm}. ORBIT addresses these problems by offering flexible reasoning modes at inference time.

\paragraph{Controllable Reasoning Generation}

Prior work for reasoning control can be broadly categorized in two main streams (1) \textit{Hard Token Budget Constraint:}
A common approach enforces reasoning efficiency through externally specified constraints, most notably by fixing explicit token budgets that cap the length of intermediate reasoning~\cite{muennighoff2025s1simpletesttimescaling, sun2025empirical, aggarwal2025l1controllinglongreasoning, anthropic2025}, relying on an absolute tokenwise budget estimation across all inputs.
(2) \textit{Discrete-Mode Systems:} Offer intuitional discrete reasoning levels, enabling users to explicitly specify their preference for the efficiency-performance trade-off~\cite{gpt-oss,heThinkDialOpenRecipe2025}, representing a soft form of control based on model's own capability. 
ORBIT falls into this category, controlling reasoning effort relative to the task through explicit mode selection.

\section{Conclusion}

In this paper, we propose ORBIT, a fully on-policy training framework for learning controllable reasoning models that operates at a target reasoning level while maintaining high reasoning density within each mode. We employ a simple truncation reward and a one-time Expansion–Compression Loop to explore the reasoning frontier, followed by On-Policy Distillation to consolidate multiple modes into a unified model initialized from merged frontier policies, which enables inference-time mode control via natural language prompts. 
Experiments across LRMs of different sizes show that ORBIT enables controllable reasoning across modes while maintaining competitive accuracy–length trade-offs within each mode compared to existing RL-based efficient reasoning methods.

\section*{Limitations}

We elucidate the limitations of this work as follows: First, long-form RL-based training can be computationally expensive for larger models, since each rollout requires autoregressive sampling, which may limit scalability to very large architectures or long-context reasoning tasks. Moreover, when the model output entropy is low, on-policy RL training may suffer from low sample efficiency, limiting further compression progress.
Second, our evaluation mainly focuses on reasoning tasks with relatively structured verification signals, and we have not extended experiments to tasks where correctness is difficult to verify with rule-based criteria. 
Finally, despite the empirical results, we do not establish theoretical guarantees on the convergence or optimality of the proposed training procedure.

\section*{Ethical Considerations}

% \subsection*{Use of AI Assistants}

% The algorithmic design and main methodology presented in this work were developed through human-led research and reasoning. During implementation, we utilized GitHub Copilot 3\footnote{https://github.com/features/copilot} for coding assistance. We affirm that the primary content and logic of the code are entirely our own work.

\subsection*{Potential Risks}

This work introduces ORBIT (On-policy Exploration-Exploitation for Controllable Multi-Budget Reasoning), a framework for training language models with controllable reasoning modes. We acknowledge potential risks associated with controllable reasoning models. Misuse could lead to over-reliance on automated reasoning, propagation of incorrect conclusions, or reinforcement of biases present in the training data. Moreover, the ability to choose reasoning modes in ORBIT may lead to misleading outputs if used improperly or adversarially.

\bibliography{main}

\begin{thebibliography}{46}
\providecommand{\natexlab}[1]{#1}

\bibitem[{Agarwal et~al.(2023)Agarwal, Vieillard, Stanczyk, Ramos, Geist, and Bachem}]{agarwal2023gkd}
Rishabh Agarwal, Nino Vieillard, Piotr Stanczyk, Sabela Ramos, Matthieu Geist, and Olivier Bachem. 2023.
\newblock Gkd: Generalized knowledge distillation for auto-regressive sequence models.
\newblock \emph{CoRR}.

\bibitem[{Aggarwal et~al.(2025)Aggarwal, Kim, Lanchantin, Welleck, Weston, Kulikov, and Saha}]{aggarwal2025optimalthinkingbenchevaluatingunderthinkingllms}
Pranjal Aggarwal, Seungone Kim, Jack Lanchantin, Sean Welleck, Jason Weston, Ilia Kulikov, and Swarnadeep Saha. 2025.
\newblock \href {https://arxiv.org/abs/2508.13141} {Optimalthinkingbench: Evaluating over and underthinking in llms}.
\newblock \emph{Preprint}, arXiv:2508.13141.

\bibitem[{Aggarwal and Welleck(2025)}]{aggarwal2025l1controllinglongreasoning}
Pranjal Aggarwal and Sean Welleck. 2025.
\newblock \href {https://arxiv.org/abs/2503.04697} {L1: Controlling how long a reasoning model thinks with reinforcement learning}.
\newblock \emph{Preprint}, arXiv:2503.04697.

\bibitem[{An et~al.(2025)An, Xie, Li, Li, Zhang, Gong, Zhong, Xu, Qiu, Wang, and Kong}]{Polaris2025}
Chenxin An, Zhihui Xie, Xiaonan Li, Lei Li, Jun Zhang, Shansan Gong, Ming Zhong, Jingjing Xu, Xipeng Qiu, Mingxuan Wang, and Lingpeng Kong. 2025.
\newblock \href {https://hkunlp.github.io/blog/2025/Polaris} {Polaris: A post-training recipe for scaling reinforcement learning on advanced reasoning models}.

\bibitem[{Anthropic(2025)}]{anthropic2025}
Anthropic. 2025.
\newblock \href {https://docs.anthropic.com/en/docs/build-with-claude/extended-thinking} {Building with extended thinking}.

\bibitem[{Arora and Zanette(2025)}]{arora2025traininglanguagemodelsreason}
Daman Arora and Andrea Zanette. 2025.
\newblock \href {https://arxiv.org/abs/2502.04463} {Training language models to reason efficiently}.
\newblock \emph{Preprint}, arXiv:2502.04463.

\bibitem[{Bengio et~al.(2015)Bengio, Vinyals, Jaitly, and Shazeer}]{bengio2015scheduled}
Samy Bengio, Oriol Vinyals, Navdeep Jaitly, and Noam Shazeer. 2015.
\newblock Scheduled {{Sampling}} for {{Sequence Prediction}} with {{Recurrent Neural Networks}}.
\newblock In \emph{Advances in {{Neural Information Processing Systems}}}, volume~28.

\bibitem[{ByteDance-Seed(2025)}]{bytedance_seed_2025_beyondaime}
ByteDance-Seed. 2025.
\newblock Beyondaime: Advancing math reasoning evaluation beyond high school olympiads.
\newblock \url{[https://huggingface.co/datasets/ByteDance-Seed/BeyondAIME](https://huggingface.co/datasets/ByteDance-Seed/BeyondAIME)}.

\bibitem[{Chen et~al.(2025{\natexlab{a}})Chen, Xu, Liang, He, Pang, Yu, Song, Liu, Zhou, Zhang, Wang, Tu, Mi, and Yu}]{chen2025think23overthinkingo1like}
Xingyu Chen, Jiahao Xu, Tian Liang, Zhiwei He, Jianhui Pang, Dian Yu, Linfeng Song, Qiuzhi Liu, Mengfei Zhou, Zhuosheng Zhang, Rui Wang, Zhaopeng Tu, Haitao Mi, and Dong Yu. 2025{\natexlab{a}}.
\newblock \href {https://arxiv.org/abs/2412.21187} {Do not think that much for 2+3=? on the overthinking of o1-like llms}.
\newblock \emph{Preprint}, arXiv:2412.21187.

\bibitem[{Chen et~al.(2025{\natexlab{b}})Chen, Ma, Fang, Yu, and Wang}]{chen2025verithinkerlearningverifymakes}
Zigeng Chen, Xinyin Ma, Gongfan Fang, Ruonan Yu, and Xinchao Wang. 2025{\natexlab{b}}.
\newblock \href {https://arxiv.org/abs/2505.17941} {Verithinker: Learning to verify makes reasoning model efficient}.
\newblock \emph{Preprint}, arXiv:2505.17941.

\bibitem[{Fang et~al.(2025)Fang, Ma, and Wang}]{fang2025thinklessllmlearnsthink}
Gongfan Fang, Xinyin Ma, and Xinchao Wang. 2025.
\newblock \href {https://arxiv.org/abs/2505.13379} {Thinkless: Llm learns when to think}.
\newblock \emph{Preprint}, arXiv:2505.13379.

\bibitem[{Graves(2017)}]{graves2017adaptivecomputationtimerecurrent}
Alex Graves. 2017.
\newblock \href {https://arxiv.org/abs/1603.08983} {Adaptive computation time for recurrent neural networks}.
\newblock \emph{Preprint}, arXiv:1603.08983.

\bibitem[{Guo et~al.(2025)Guo, Yang, Zhang, Song, Wang, Zhu, Xu, Zhang, Ma, Bi, Zhang, Yu, Wu, Wu, Gou, Shao, Li, Gao, Liu, Xue, Wang, Wu, Feng, Lu, Zhao, Deng, Ruan, Dai, Chen, Ji, Li, Lin, Dai, Luo, Hao, Chen, Li, Zhang, Xu, Ding, Gao, Qu, Li, Guo, Li, Chen, Yuan, Tu, Qiu, Li, Cai, Ni, Liang, Chen, Dong, Hu, You, Gao, Guan, Huang, Yu, Wang, Zhang, Zhao, Wang, Zhang, Xu, Xia, Zhang, Zhang, Tang, Zhou, Li, Wang, Li, Tian, Huang, Zhang, Wang, Chen, Du, Ge, Zhang, Pan, Wang, Chen, Jin, Chen, Lu, Zhou, Chen, Ye, Wang, Yu, Zhou, Pan, Li, Zhou, Wu, Yun, Pei, Sun, Wang, Zeng, Liu, Liang, Gao, Yu, Zhang, Xiao, An, Liu, Wang, Chen, Nie, Cheng, Liu, Xie, Liu, Yang, Li, Su, Lin, Li, Jin, Shen, Chen, Sun, Wang, Song, Zhou, Wang, Shan, Li, Wang, Wei, Zhang, Xu, Li, Zhao, Sun, Wang, Yu, Zhang, Shi, Xiong, He, Piao, Wang, Tan, Ma, Liu, Guo, Ou, Wang, Gong, Zou, He, Xiong, Luo, You, Liu, Zhou, Zhu, Huang, Li, Zheng, Zhu, Ma, Tang, Zha, Yan, Ren, Ren, Sha, Fu, Xu, Xie, Zhang, Hao, Ma, Yan, Wu, Gu, Zhu, Liu, Li, Xie, Song,
  Pan, Huang, Xu, Zhang, and Zhang}]{guo2025deepseekr1}
Daya Guo, Dejian Yang, Haowei Zhang, Junxiao Song, Peiyi Wang, Qihao Zhu, Runxin Xu, Ruoyu Zhang, Shirong Ma, Xiao Bi, Xiaokang Zhang, Xingkai Yu, Yu~Wu, Z.~F. Wu, Zhibin Gou, Zhihong Shao, Zhuoshu Li, Ziyi Gao, Aixin Liu, and 175 others. 2025.
\newblock {{DeepSeek-R1}} incentivizes reasoning in {{LLMs}} through reinforcement learning.
\newblock \emph{Nature}, 645:633--638.

\bibitem[{He et~al.(2025)He, Yuan, Li, Wang, and Chen}]{heThinkDialOpenRecipe2025}
Qianyu He, Siyu Yuan, Xuefeng Li, Mingxuan Wang, and Jiangjie Chen. 2025.
\newblock \href {https://arxiv.org/abs/2508.18773} {{{ThinkDial}}: {{An Open Recipe}} for {{Controlling Reasoning Effort}} in {{Large Language Models}}}.
\newblock \emph{Preprint}, arXiv:2508.18773.

\bibitem[{Jang et~al.(2026)Jang, Yeom, Yeo, Lim, and Kim}]{jang2026stable}
Ijun Jang, Jewon Yeom, Juan Yeo, Hyunggu Lim, and Taesup Kim. 2026.
\newblock \href {https://arxiv.org/abs/2601.07155} {Stable {{On-Policy Distillation}} through {{Adaptive Target Reformulation}}}.
\newblock \emph{Preprint}, arXiv:2601.07155.

\bibitem[{Jiang et~al.(2025)Jiang, Quan, Ding, Luo, Wang, and Hu}]{jiang2025flashthinkearlyexitmethod}
Guochao Jiang, Guofeng Quan, Zepeng Ding, Ziqin Luo, Dixuan Wang, and Zheng Hu. 2025.
\newblock \href {https://arxiv.org/abs/2505.13949} {Flashthink: An early exit method for efficient reasoning}.
\newblock \emph{Preprint}, arXiv:2505.13949.

\bibitem[{Jiao et~al.(2025)Jiao, Gao, Yang, Zhou, Huang, Chen, and Li}]{jiao-etal-2025-analyzing}
Liuxuan Jiao, Chen Gao, Yiqian Yang, Chenliang Zhou, YiXian Huang, Xinlei Chen, and Yong Li. 2025.
\newblock \href {https://doi.org/10.18653/v1/2025.emnlp-main.1676} {Analyzing and modeling {LLM} response lengths with extreme value theory: Anchoring effects and hybrid distributions}.
\newblock In \emph{Proceedings of the 2025 Conference on Empirical Methods in Natural Language Processing}. Association for Computational Linguistics.

\bibitem[{Jin et~al.(2025)Jin, Yeom, Bae, and Kim}]{jin2025wellthinkingenhancingllm}
Hyunbin Jin, Je~Won Yeom, Seunghyun Bae, and Taesup Kim. 2025.
\newblock \href {https://arxiv.org/abs/2503.10167} {"well, keep thinking": Enhancing llm reasoning with adaptive injection decoding}.
\newblock \emph{Preprint}, arXiv:2503.10167.

\bibitem[{Li and Vitnyi(2008)}]{10.5555/1478784}
Ming Li and Paul~M.B. Vitnyi. 2008.
\newblock \emph{An Introduction to Kolmogorov Complexity and Its Applications}, 3 edition.
\newblock Springer Publishing Company, Incorporated.

\bibitem[{Lightman et~al.(2023)Lightman, Kosaraju, Burda, Edwards, Baker, Lee, Leike, Schulman, Sutskever, and Cobbe}]{math500}
Hunter Lightman, Vineet Kosaraju, Yura Burda, Harri Edwards, Bowen Baker, Teddy Lee, Jan Leike, John Schulman, Ilya Sutskever, and Karl Cobbe. 2023.
\newblock \href {https://arxiv.org/abs/2305.20050} {Let's verify step by step}.
\newblock \emph{Preprint}, arXiv:2305.20050.

\bibitem[{Liu et~al.(2025)Liu, Zhou, Deng, Huang, Liu, Deng, Zhang, and He}]{liu2025learnreasonefficientlyadaptive}
Wei Liu, Ruochen Zhou, Yiyun Deng, Yuzhen Huang, Junteng Liu, Yuntian Deng, Yizhe Zhang, and Junxian He. 2025.
\newblock \href {https://arxiv.org/abs/2505.15612} {Learn to reason efficiently with adaptive length-based reward shaping}.
\newblock \emph{Preprint}, arXiv:2505.15612.

\bibitem[{Lou et~al.(2025)Lou, Sun, Liang, Qu, Shen, Wang, Li, Yang, and Wu}]{lou2025adacot}
Chenwei Lou, Zewei Sun, Xinnian Liang, Meng Qu, Wei Shen, Wenqi Wang, Yuntao Li, Qingping Yang, and Shuangzhi Wu. 2025.
\newblock Adacot: Pareto-optimal adaptive chain-of-thought triggering via reinforcement learning.
\newblock \emph{arXiv preprint arXiv:2505.11896}.

\bibitem[{Lu and Lab(2025)}]{lu2025onpolicydistillation}
Kevin Lu and Thinking~Machines Lab. 2025.
\newblock \href {https://doi.org/10.64434/tml.20251026} {On-policy distillation}.
\newblock \emph{Thinking Machines Lab: Connectionism}.
\newblock Https://thinkingmachines.ai/blog/on-policy-distillation.

\bibitem[{Luo et~al.(2025{\natexlab{a}})Luo, He, Wang, Yang, Liu, Tan, Cao, Tao, and Shen}]{luo2025adar1hybridcotbileveladaptive}
Haotian Luo, Haiying He, Yibo Wang, Jinluan Yang, Rui Liu, Naiqiang Tan, Xiaochun Cao, Dacheng Tao, and Li~Shen. 2025{\natexlab{a}}.
\newblock \href {https://arxiv.org/abs/2504.21659} {Ada-r1: Hybrid-cot via bi-level adaptive reasoning optimization}.
\newblock \emph{Preprint}, arXiv:2504.21659.

\bibitem[{Luo et~al.(2025{\natexlab{b}})Luo, Shen, He, Wang, Liu, Li, Tan, Cao, and Tao}]{luo2025o1prunerlengthharmonizingfinetuningo1like}
Haotian Luo, Li~Shen, Haiying He, Yibo Wang, Shiwei Liu, Wei Li, Naiqiang Tan, Xiaochun Cao, and Dacheng Tao. 2025{\natexlab{b}}.
\newblock \href {https://arxiv.org/abs/2501.12570} {O1-pruner: Length-harmonizing fine-tuning for o1-like reasoning pruning}.
\newblock \emph{Preprint}, arXiv:2501.12570.

\bibitem[{Muennighoff et~al.(2025)Muennighoff, Yang, Shi, Li, Fei-Fei, Hajishirzi, Zettlemoyer, Liang, Candès, and Hashimoto}]{muennighoff2025s1simpletesttimescaling}
Niklas Muennighoff, Zitong Yang, Weijia Shi, Xiang~Lisa Li, Li~Fei-Fei, Hannaneh Hajishirzi, Luke Zettlemoyer, Percy Liang, Emmanuel Candès, and Tatsunori Hashimoto. 2025.
\newblock \href {https://arxiv.org/abs/2501.19393} {s1: Simple test-time scaling}.
\newblock \emph{Preprint}, arXiv:2501.19393.

\bibitem[{OpenAI(2025)}]{gpt-oss}
OpenAI. 2025.
\newblock \href {https://cdn.openai.com/pdf/419b6906-9da6-406c-a19d-1bb078ac7637/oai_gpt-oss_model_card.pdf} {gpt-oss-120b \& gpt-oss-20b model card}.

\bibitem[{{OpenAI}(2025)}]{o3mini}
{OpenAI}. 2025.
\newblock \href {https://cdn.openai.com/o3-mini-system-card-feb10.pdf} {Openai o3-mini system card}.

\bibitem[{Rein et~al.(2024)Rein, Hou, Stickland, Petty, Pang, Dirani, Michael, and Bowman}]{gpqa}
David Rein, Betty~Li Hou, Asa~Cooper Stickland, Jackson Petty, Richard~Yuanzhe Pang, Julien Dirani, Julian Michael, and Samuel~R Bowman. 2024.
\newblock Gpqa: A graduate-level google-proof q\&a benchmark.
\newblock In \emph{First Conference on Language Modeling}.

\bibitem[{Schulman et~al.(2017)Schulman, Wolski, Dhariwal, Radford, and Klimov}]{schulman2017proximal}
John Schulman, Filip Wolski, Prafulla Dhariwal, Alec Radford, and Oleg Klimov. 2017.
\newblock Proximal policy optimization algorithms.
\newblock \emph{arXiv preprint arXiv:1707.06347}.

\bibitem[{Setlur et~al.(2025)Setlur, Yang, Snell, Greer, Wu, Smith, Simchowitz, and Kumar}]{setlur2025e3}
Amrith Setlur, Matthew~YR Yang, Charlie Snell, Jeremy Greer, Ian Wu, Virginia Smith, Max Simchowitz, and Aviral Kumar. 2025.
\newblock e3: Learning to explore enables extrapolation of test-time compute for llms.
\newblock \emph{arXiv preprint arXiv:2506.09026}.

\bibitem[{Shao et~al.(2024)Shao, Wang, Zhu, Xu, Song, Bi, Zhang, Zhang, Li, Wu et~al.}]{shao2024deepseekmath}
Zhihong Shao, Peiyi Wang, Qihao Zhu, Runxin Xu, Junxiao Song, Xiao Bi, Haowei Zhang, Mingchuan Zhang, YK~Li, Yang Wu, and 1 others. 2024.
\newblock Deepseekmath: Pushing the limits of mathematical reasoning in open language models.
\newblock \emph{arXiv preprint arXiv:2402.03300}.

\bibitem[{Sheng et~al.(2024)Sheng, Zhang, Ye, Wu, Zhang, Zhang, Peng, Lin, and Wu}]{sheng2024hybridflow}
Guangming Sheng, Chi Zhang, Zilingfeng Ye, Xibin Wu, Wang Zhang, Ru~Zhang, Yanghua Peng, Haibin Lin, and Chuan Wu. 2024.
\newblock Hybridflow: A flexible and efficient rlhf framework.
\newblock \emph{arXiv preprint arXiv: 2409.19256}.

\bibitem[{Sun et~al.(2025)Sun, Wang, Li, Liu, Li, Wen, Yuan, Zheng, Liang, Li et~al.}]{sun2025empirical}
Yi~Sun, Han Wang, Jiaqiang Li, Jiacheng Liu, Xiangyu Li, Hao Wen, Yizhen Yuan, Huiwen Zheng, Yan Liang, Yuanchun Li, and 1 others. 2025.
\newblock An empirical study of llm reasoning ability under strict output length constraint.
\newblock \emph{arXiv preprint arXiv:2504.14350}.

\bibitem[{Team et~al.(2025)Team, Du, Gao, Xing, Jiang, Chen, Li, Xiao, Du, Liao, Tang, Wang, Zhang, Yuan, Lu, Tang, Sung, Wei, Lai, Guo, Zhu, Ding, Hu, Yang, Zhang, Yao, Zhao, Lu, Li, Yu, Gao, Zheng, Yuan, Chen, Guo, Su, Wang, Zhao, Zhang, Liu, Yan, Wu, Shi, Ye, Yu, Dong, Zhang, Ma, Pan, Gong, Liu, Ma, Wei, Cao, Huang, Jiang, Gao, Xiong, He, Huang, Xu, Wu, He, Wei, Jia, Wu, Xu, Zu, Zhou, Pan, Charles, Li, Hu, Liu, Chen, Wang, Liu, Qin, Liu, Yang, Bao, Du, Wu, Wang, Zhou, Wang, Li, Zhu, Zhang, Wang, Yang, Huang, Huang, Xu, Yang, and Lin}]{team2025kimi}
Kimi Team, Angang Du, Bofei Gao, Bowei Xing, Changjiu Jiang, Cheng Chen, Cheng Li, Chenjun Xiao, Chenzhuang Du, Chonghua Liao, Chuning Tang, Congcong Wang, Dehao Zhang, Enming Yuan, Enzhe Lu, Fengxiang Tang, Flood Sung, Guangda Wei, Guokun Lai, and 77 others. 2025.
\newblock \href {https://arxiv.org/abs/2501.12599} {Kimi k1.5: {{Scaling Reinforcement Learning}} with {{LLMs}}}.
\newblock \emph{Preprint}, arXiv:2501.12599.

\bibitem[{Tu et~al.(2025)Tu, Lin, Zhang, Tian, Li, Lan, and Zhao}]{tu2025learningthinkshapingadaptive}
Songjun Tu, Jiahao Lin, Qichao Zhang, Xiangyu Tian, Linjing Li, Xiangyuan Lan, and Dongbin Zhao. 2025.
\newblock \href {https://arxiv.org/abs/2505.10832} {Learning when to think: Shaping adaptive reasoning in r1-style models via multi-stage rl}.
\newblock \emph{Preprint}, arXiv:2505.10832.

\bibitem[{Wang et~al.(2025)Wang, Feng, Chen, Chu, Krishna, and Zhou}]{wang2025waitdontneedwait}
Chenlong Wang, Yuanning Feng, Dongping Chen, Zhaoyang Chu, Ranjay Krishna, and Tianyi Zhou. 2025.
\newblock \href {https://arxiv.org/abs/2506.08343} {Wait, we don't need to "wait"! removing thinking tokens improves reasoning efficiency}.
\newblock \emph{Preprint}, arXiv:2506.08343.

\bibitem[{Wang et~al.(2024)Wang, Ma, Zhang, Ni, Chandra, Guo, Ren, Arulraj, He, Jiang et~al.}]{mmlu-pro}
Yubo Wang, Xueguang Ma, Ge~Zhang, Yuansheng Ni, Abhranil Chandra, Shiguang Guo, Weiming Ren, Aaran Arulraj, Xuan He, Ziyan Jiang, and 1 others. 2024.
\newblock Mmlu-pro: A more robust and challenging multi-task language understanding benchmark.
\newblock \emph{Advances in Neural Information Processing Systems}, 37:95266--95290.

\bibitem[{Wei et~al.(2022)Wei, Wang, Schuurmans, Bosma, Ichter, Xia, Chi, Le, and Zhou}]{weichainofthought}
Jason Wei, Xuezhi Wang, Dale Schuurmans, Maarten Bosma, Brian Ichter, Fei Xia, Ed~H Chi, Quoc~V Le, and Denny Zhou. 2022.
\newblock Chain-of-{{Thought Prompting Elicits Reasoning}} in {{Large Language Models}}.

\bibitem[{Wortsman et~al.(2022)Wortsman, Ilharco, Gadre, Roelofs, Gontijo-Lopes, Morcos, Namkoong, Farhadi, Carmon, Kornblith, and Schmidt}]{wortsman2022modelsoupsaveragingweights}
Mitchell Wortsman, Gabriel Ilharco, Samir~Yitzhak Gadre, Rebecca Roelofs, Raphael Gontijo-Lopes, Ari~S. Morcos, Hongseok Namkoong, Ali Farhadi, Yair Carmon, Simon Kornblith, and Ludwig Schmidt. 2022.
\newblock \href {https://arxiv.org/abs/2203.05482} {Model soups: averaging weights of multiple fine-tuned models improves accuracy without increasing inference time}.
\newblock \emph{Preprint}, arXiv:2203.05482.

\bibitem[{Xiaomi et~al.(2025)Xiaomi, Xia, Shen, Cici, Zhu, Zhang, Wang, Zhang, Liu, Xiao, Dong, Zhao, Li, Wang, Yu, Chen, Wang, Ma, Deng, Huang, Song, Jiang, Ye, Cai, He, Zhang, Zhang, Wang, Tian, Zhao, Qu, Xu, Shi, Bao, Fang, Zhou, Zhou, Li, Zhu, Chen, Wang, Liu, Li, Gu, Ren, Liu, Deng, Zhuang, Lv, Yang, Zhang, Yong, Zhang, Song, Xu, Wang, Yan, Tu, Tian, Wang, Yu, Lin, Song, and Yue}]{xiaomi2025mimo}
LLM-Core Xiaomi, Bingquan Xia, Bowen Shen, Cici, Dawei Zhu, Di~Zhang, Gang Wang, Hailin Zhang, Huaqiu Liu, Jiebao Xiao, Jinhao Dong, Liang Zhao, Peidian Li, Peng Wang, Shihua Yu, Shimao Chen, Weikun Wang, Wenhan Ma, Xiangwei Deng, and 45 others. 2025.
\newblock \href {https://arxiv.org/abs/2505.07608} {{{MiMo}}: {{Unlocking}} the {{Reasoning Potential}} of {{Language Model}} -- {{From Pretraining}} to {{Posttraining}}}.
\newblock \emph{Preprint}, arXiv:2505.07608.

\bibitem[{Yang et~al.(2025{\natexlab{a}})Yang, Si, Duan, Zhu, Zhu, Li, Lin, Cao, and Wang}]{yang2025dynamicearlyexitreasoning}
Chenxu Yang, Qingyi Si, Yongjie Duan, Zheliang Zhu, Chenyu Zhu, Qiaowei Li, Zheng Lin, Li~Cao, and Weiping Wang. 2025{\natexlab{a}}.
\newblock \href {https://arxiv.org/abs/2504.15895} {Dynamic early exit in reasoning models}.
\newblock \emph{Preprint}, arXiv:2504.15895.

\bibitem[{Yang et~al.(2025{\natexlab{b}})Yang, Ma, Lin, and Wei}]{yang2025thinking}
Wenkai Yang, Shuming Ma, Yankai Lin, and Furu Wei. 2025{\natexlab{b}}.
\newblock \href {https://arxiv.org/abs/2502.18080} {Towards thinking-optimal scaling of test-time compute for llm reasoning}.
\newblock \emph{Preprint}, arXiv:2502.18080.

\bibitem[{Yi et~al.(2025)Yi, Wang, and Li}]{yi2025shorterbetter}
Jingyang Yi, Jiazheng Wang, and Sida Li. 2025.
\newblock Shorterbetter: Guiding reasoning models to find optimal inference length for efficient reasoning.
\newblock \emph{arXiv preprint arXiv:2504.21370}.

\bibitem[{Yu et~al.(2025)Yu, Zhang, Zhu, Yuan, Zuo, Yue, Dai, Fan, Liu, Liu, Liu, Lin, Lin, Ma, Sheng, Tong, Zhang, Zhang, Zhang, Zhu, Zhu, Chen, Chen, Wang, Yu, Song, Wei, Zhou, Liu, Ma, Zhang, Yan, Qiao, Wu, and Wang}]{yu2025dapo}
Qiying Yu, Zheng Zhang, Ruofei Zhu, Yufeng Yuan, Xiaochen Zuo, Yu~Yue, Weinan Dai, Tiantian Fan, Gaohong Liu, Lingjun Liu, Xin Liu, Haibin Lin, Zhiqi Lin, Bole Ma, Guangming Sheng, Yuxuan Tong, Chi Zhang, Mofan Zhang, Wang Zhang, and 16 others. 2025.
\newblock \href {https://doi.org/10.48550/arXiv.2503.14476} {{{DAPO}}: {{An Open-Source LLM Reinforcement Learning System}} at {{Scale}}}.
\newblock \emph{Preprint}, arXiv:2503.14476.

\bibitem[{Zhang et~al.(2025)Zhang, Lin, Hou, Feng, and Li}]{zhang2025adaptthinkreasoningmodelslearn}
Jiajie Zhang, Nianyi Lin, Lei Hou, Ling Feng, and Juanzi Li. 2025.
\newblock \href {https://arxiv.org/abs/2505.13417} {Adaptthink: Reasoning models can learn when to think}.
\newblock \emph{Preprint}, arXiv:2505.13417.

\end{thebibliography}
% \clearpage

\appendix
\label{sec:appendix}
\section{Intuition for Logarithmic Compression Scheduling}
\label{app:discussion}
The logarithmic schedule used in the expansion--compression loop
can be motivated by a simple structural property under heavy-tailed
trajectory length distributions in LLM responses~\citep{jiao-etal-2025-analyzing}.

Let $T:=|o|$ denote the trajectory length.
Define the tail probability
\[
S(L) := \mathbb{P}(T > L),
\]
which measures the probability mass of trajectories
whose length exceeds the context budget $L$.

Over the relevant range of budgets, we approximate the tail distribution by a heavy-tailed form:

\begin{equation}
S(L) = C L^{-\alpha}, \qquad C>0,\ \alpha>0 .
\end{equation}

Under a multiplicative compression schedule
\begin{equation}
L_{k+1} = \rho L_k , \qquad 0<\rho<1 ,
\end{equation}
the truncated trajectory mass satisfies
\begin{align}
\frac{S(L_{k+1})}{S(L_k)}
&=
\frac{C(\rho L_k)^{-\alpha}}{C L_k^{-\alpha}}
=
\rho^{-\alpha}.
\end{align}
which is independent of $k$.
Thus each compression stage imposes a constant relative truncation pressure on the trajectory distribution.

By contrast, for example, under an additive schedule
\begin{equation}
L_{k+1} = L_k - \Delta , \qquad \Delta>0 ,
\end{equation}
the corresponding change becomes
\begin{align}
\frac{S(L_{k+1})}{S(L_k)}
=
\left(\frac{L_k}{L_k-\Delta}\right)^{\alpha},
\end{align}
which depends on $L_k$ and grows as $L_k$ decreases.

We therefore take a logarithmic scheduling with $\rho=0.5$ in ORBIT for a balance between stable truncation and clear separation of different reasoning policies. 
\section{Mode-aware initialization via Model Merging }
\label{app:model_merge}

As described in Sec.~\ref{sec:exploitation}, optimizing Eq. \ref{equ:okd} in practice faces a "cold-start" challenge. To mitigate this misalignment, we proposed using model merging as an initialization of the student model.

Model merging acts as a simple linear combination of model parameters. Numerous methods have been proposed, we use average merging~\citep{wortsman2022modelsoupsaveragingweights} here for simplicity.
This process can be defined as follows:
\begin{equation}
    \phi = \frac{1}{K} \sum_{k=1}^{K} \theta_k
\end{equation}

We compare two initialization strategies: the merged model and the final-stage policy from sequential multi-stage RL, as these are the only models trained under all reasoning modes.

We first use the Accumulated Misalignment Rate (AMR), which measures whether the top-1 token probabilities differ between two models. This metric aligns naturally with the on-policy distillation process and provides an intuitive visualization of behavioral differences. Specifically, let $a_t = \arg\max \pi(o_t|q, o_{<t})$ and $\hat{a}_t = \arg\max \pi_{\theta_k}(o_t|q, o_{<t})$. The AMR is defined as:
\begin{equation}
\begin{split}
    \text{AMR}(\pi, \pi_{\theta_{k}}) &= \mathbb{E}_{q \sim \mathcal{D}, o \sim \pi(\cdot|q)} \left[ \frac{1}{|o|} \sum_{t=1}^{|o|} \mathbb{I}(a_t \neq \hat{a}_t) \right]
\end{split}
\end{equation}

\begin{figure}[!t]
    \includegraphics[width=\columnwidth]{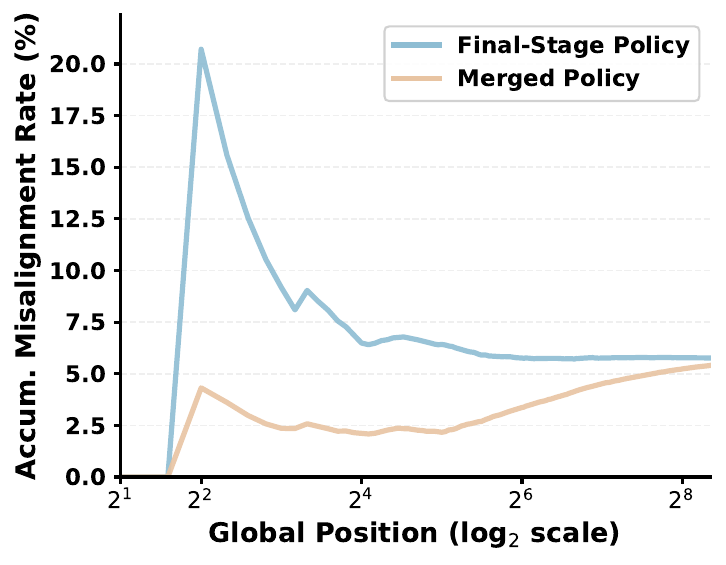}
    \caption{Misalignment rate on sequences generated by the merged model and the low-mode model.}
    \label{fig:misalignment_rate}
\end{figure}

\begin{figure}[!t]
    \includegraphics[width=\columnwidth]{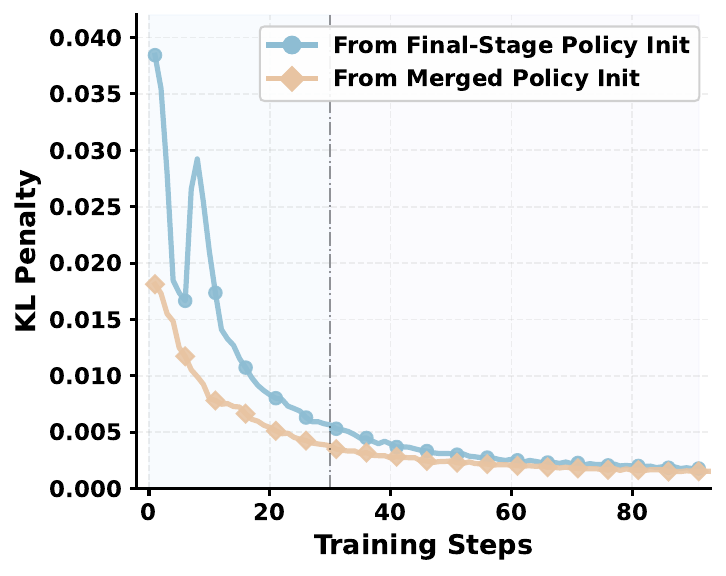}
    \caption{RKL training loss dynamics under different initializations in OPD. 30 steps denotes convergence.}
    \label{fig:merge_ablation}
\end{figure}
As shown in Figure~\ref{fig:misalignment_rate}, this merged  initialization significantly reduces the initial misalignment gap, providing a more stable starting point and accelerating the convergence of subsequent RKL distillation.

We further examine the training performance under different initialization, as shown in Figure~\ref{fig:merge_ablation}, merged model consistently outperforms final stage policy by a lower per token RKL penalty, demonstrating the effectiveness of such initialization.

\section{Training Details}
\label{app:train_detail}

We build on the VeRL codebase~\citep{sheng2024hybridflow} with RLVR. Following the DAPO recipe~\citep{yu2025dapo}, which allows for higher clipping—a practice that has been widely adopted in RL training, we also use rejection sampling as a further ensurance for training stability.

\textbf{Training Configuration.}  
We maintain consistent hyperparameters throughout the exploration RL stage, using a batch size of $256$, rejection pools of $512$, and a learning rate of $1 \times 10^{-6}$ without warm-up. For generation, we apply standard rollout settings: temperature $1$, \texttt{top-p} $1$, \texttt{top-k} $-1$, and $8$ rollouts per problem. 
We train \texttt{DeepSeek-Distill-Qwen-1.5B} using the DAPO-Math-17K dataset~\citep{yu2025dapo}, while \texttt{Qwen3-4B-2507-Thinking} and \texttt{Nemotron-7B} are trained on the Polaris-53K~\citep{Polaris2025}.

For exploitation, we train OPD with a batch size of $1024$, single generation per prompt, and a token-wise KL penalty coefficient of $1.0$.  

\textbf{Context Window Scheduling.}
We implement an expansion-compression loop strategy to dynamically manage context windows in Exploration stage RL. Specifically, the maximum output length follows a schedule of $32\text{K} \rightarrow 16\text{K} \rightarrow 8\text{K} \rightarrow 4\text{K}$ for \texttt{Qwen3-4B-2507-Thinking} and \texttt{Openmath-Nemotron-7B}, and $16\text{K} \rightarrow 8\text{K} \rightarrow 4\text{K} \rightarrow 2\text{K}$ for \texttt{DeepSeek-Distill-Qwen-1.5B}. In both cases, the first stage of the context window serves as the expansion phase.
\section{System Prompts For Each Reasoning Mode}

To build a semantically meaningful guidance for discrete mode separated reasoning behavior, we follow \cite{yang2025thinking} and generalized it to four modes, where each one of them represents semactically aligned expected reasoning behavior under corresponding mode.

\begin{table*}
\resizebox{\linewidth}{!}{
\begin{tcolorbox}
\textit{\color{gray}{/* Low Mode */}} \\
You have extremely limited time to think and respond to the user's query. Every additional second of processing and reasoning incurs a significant resource cost, which could affect efficiency and effectiveness. Your task is to prioritize speed without sacrificing essential clarity or accuracy. Provide the most direct and concise answer possible. Avoid unnecessary steps, reflections, verification, or refinements UNLESS ABSOLUTELY NECESSARY. Your primary goal is to deliver a quick, clear and correct response.

\vspace{1mm}

\textit{\color{gray}{/* Mid Mode */}} \\
You have sufficient time to think and respond to the user's query, allowing for a more thoughtful and in-depth answer. However, be aware that the longer you take to reason and process, the greater the associated resource costs and potential consequences. While you should not rush, aim to balance the depth of your reasoning with efficiency. Prioritize providing a well-thought-out response, but do not overextend your thinking if the answer can be provided with a reasonable level of analysis. Use your reasoning time wisely, focusing on what is essential for delivering an accurate response without unnecessary delays and overthinking.

\vspace{1mm}

\textit{\color{gray}{/* High Mode */}} \\
You have a substantial amount of time to think and respond. The pressure on reasoning cost is low, allowing you to prioritize accuracy and robustness over speed. You should verify your key steps and consider alternative interpretations to ensure your answer is reliable. While you are encouraged to think deeply, aim for a confident solution rather than an endless exploration. If you encounter ambiguity, take the time to resolve it, but remain focused on deriving the correct conclusion efficiently within a generous budget.

\vspace{1mm}

\textit{\color{gray}{/* Xhigh Mode */}} \\
You have unlimited time to think and respond. There are absolutely no constraints on reasoning length or associated costs. Your singular goal is to achieve the highest possible level of accuracy and certainty. You must exhaustively explore the problem space, considering all edge cases, theoretical nuances, and alternative methodologies. Engage in rigorous, iterative self-correction and deep verification to eliminate even the slightest possibility of error. Do not settle for a 'likely' correct answer; explore every angle until you are certain. Perfection is the only standard.

\end{tcolorbox}
}
\caption{System prompts used for triggering Low, Mid, High, Xhigh modes}
\label{tab:sys_prompts}

\end{table*}

\section{Additional Evaluation Results}

\newcommand{\teacher}{Teachers}

\definecolor{lightblue1}{RGB}{240, 248, 255}
\definecolor{lightblue2}{RGB}{220, 240, 255}
\definecolor{lightblue3}{RGB}{200, 230, 255}
\definecolor{lightblue4}{RGB}{180, 220, 255}
\definecolor{lightblue5}{RGB}{160, 210, 255}

\definecolor{lightgreen1}{RGB}{240, 255, 240}
\definecolor{lightgreen2}{RGB}{220, 245, 220}
\definecolor{lightgreen3}{RGB}{200, 235, 200}
\definecolor{lightgreen4}{RGB}{180, 225, 180}
\definecolor{lightgreen5}{RGB}{160, 215, 160}

\definecolor{lightcyan1}{RGB}{240, 255, 255}
\definecolor{lightcyan2}{RGB}{220, 250, 250}
\definecolor{lightcyan3}{RGB}{200, 245, 245}
\definecolor{lightcyan4}{RGB}{180, 240, 240}
\definecolor{lightcyan5}{RGB}{160, 235, 235}

\definecolor{lightpurple1}{RGB}{245, 240, 255}
\definecolor{lightpurple2}{RGB}{235, 220, 255}
\definecolor{lightpurple3}{RGB}{225, 200, 255}
\definecolor{lightpurple4}{RGB}{215, 180, 255}
\definecolor{lightpurple5}{RGB}{205, 160, 255}
\definecolor{lightpurple6}{RGB}{195, 140, 255}

\definecolor{lightred1}{RGB}{255, 240, 240}
\definecolor{lightred2}{RGB}{255, 220, 220}
\definecolor{lightred3}{RGB}{255, 200, 200}
\definecolor{lightred4}{RGB}{255, 180, 180}
\definecolor{lightred5}{RGB}{255, 160, 160}
\definecolor{lightred6}{RGB}{255, 140, 140}

\definecolor{lightorange1}{RGB}{255, 245, 230}
\definecolor{lightorange2}{RGB}{255, 235, 210}
\definecolor{lightorange3}{RGB}{255, 225, 190}
\definecolor{lightorange4}{RGB}{255, 215, 170}
\definecolor{lightorange5}{RGB}{255, 205, 150}
\definecolor{lightorange6}{RGB}{255, 195, 130}

\definecolor{lightyellow1}{RGB}{255, 255, 240}
\definecolor{lightyellow2}{RGB}{255, 250, 210}
\definecolor{lightyellow3}{RGB}{255, 245, 180}
\definecolor{lightyellow4}{RGB}{255, 240, 150}

\begin{table*}[!t]
    \centering
    \vspace{2mm}
    \resizebox{0.95\textwidth}{!}{
    \begin{tabular}{l|cc|cc|cc|cc|cc|cc}
    		    \toprule[1.6pt]
        & \multicolumn{6}{c|}{\textbf{Mathematics}} & \multicolumn{4}{c|}{\textbf{Multi-Task}} & \multicolumn{2}{c}{\multirow{2}{*}{\textbf{Overall}}} \\
        \cmidrule(lr){2-11}
        \multirow{3}{*}{\textbf{Models}} 
           & \multicolumn{2}{c|}{\textbf{AIME 24}} 
           & \multicolumn{2}{c|}{\textbf{AIME 25}} 
           & \multicolumn{2}{c|}{\textbf{Beyond AIME}} 
           & \multicolumn{2}{c|}{\textbf{GPQA}} 
           & \multicolumn{2}{c|}{\textbf{MMLU}} 
           & \multicolumn{2}{c}{} \\ \cmidrule(lr){2-13}
        & \texttt{avg@32} & \texttt{Toks}
        & \texttt{avg@32} & \texttt{Toks}
        & \texttt{avg@8} & \texttt{Toks}
        & \texttt{avg@8} & \texttt{Toks}
        & \texttt{avg@1} & \texttt{Toks}
        & \texttt{Avg.} & \texttt{Toks} \\
        \midrule
        \multicolumn{13}{c}{\texttt{\textbf{{DeepSeek-Distill-Qwen-1.5B}}}} \\
        \midrule
        Initial Model & 28.6 & 18.8 & 22.8 & 18.5 & 10.1 & 18.6 & 28.0 & 12.7 & 35.5 & 7.2 & 25.0 & 15.2 \\
        - {\teacher} Low & 28.3 & 2.6 & 19.0 & 2.0 & 11.0 & 2.0 & 34.3 & 1.8 & 37.6 & 1.4 & 26.0 & 2.0 \\
        - {\teacher} Mid & 35.2 & 4.9 & 26.4 & 4.5 & 10.4 & 3.6 & 32.9 & 2.9 & 38.2 & 2.2 & 28.6 & 3.6 \\
        - {\teacher} High & 40.0 & 7.0 & 29.8 & 6.7 & 14.0 & 6.3 & 39.8 & 4.8 & 40.7 & 3.8 & 32.9 & 5.7 \\
        - {\teacher} Xhigh & 42.5 & 10.7 & 32.9 & 11.0 & 16.0 & 11.7 & 39.5 & 8.3 & 40.0 & 7.7 & 34.2 & 9.9 \\
        \midrule
        \multicolumn{13}{c}{\texttt{\textbf{{Qwen3-4B-2507-Thinking}}}} \\
        \midrule
        Initial Model & 81.6 & 20.1 & 77.1 & 18.3 & 52.4 & 27.6 & 66.1 & 8.8 & 74.4 & 4.4 & 70.3 & 15.8 \\
        - {\teacher} Low & 64.4 & 5.0 & 54.3 & 5.5 & 36.6 & 5.2 & 57.3 & 1.9 & 71.5 & 0.9 & 56.8 & 3.7 \\
        - {\teacher} Mid & 66.3 & 5.9 & 57.7 & 6.4 & 40.0 & 6.4 & 63.8 & 3.1 & 72.3 & 1.9 & 60.0 & 4.7 \\
        - {\teacher} High & 75.1 & 9.4 & 69.1 & 10.6 & 46.0 & 11.7 & 65.4 & 4.9 & 73.4 & 3.0 & 65.8 & 7.9 \\
        - {\teacher} Xhigh & 83.2 & 16.8 & 78.2 & 19.0 & 52.7 & 22.7 & 67.1 & 7.9 & 74.1 & 5.8 & 71.1 & 14.4 \\
        \midrule
        \multicolumn{13}{c}{\texttt{\textbf{{Openmath-Nemotron-7B}}}} \\
        \midrule
        Initial Model & 71.6 & 11.9 & 59.9 & 13.7 & 41.8 & 13.6 & 32.5 & 8.5 & 42.4 & 5.7 & 49.6 & 10.7 \\
        - {\teacher} Low & 55.2 & 4.5 & 33.1 & 4.5 & 29.8 & 3.7 & 29.3 & 2.2 & 39.2 & 1.7 & 37.3 & 3.3 \\
        - {\teacher} Mid & 67.1 & 6.7 & 44.5 & 7.6 & 36.3 & 6.6 & 32.4 & 4.4 & 41.2 & 3.0 & 44.3 & 5.7 \\
        - {\teacher} High & 74.3 & 11.1 & 62.1 & 12.9 & 41.9 & 12.1 & 30.9 & 7.9 & 42.6 & 5.3 & 50.4 & 9.9 \\
        - {\teacher} Xhigh & 75.9 & 13.2 & 65.1 & 14.9 & 43.8 & 15.0 & 33.7 & 9.8 & 43.1 & 6.5 & 52.3 & 11.9 \\
        \bottomrule[1.6pt]
    \end{tabular}}
     \caption{Comparison of reasoning modes of {\teacher} across various reasoning tasks.}
    \label{tab:teacher_overview}
%\vspace{-0.5em}
\end{table*}

\begin{table*}[!t]
    \centering
    \resizebox{0.95\textwidth}{!}{
    \begin{tabular}{l|cc|cc|cc|cc|cc|cc}
        \toprule[1.6pt]
        \multirow{2}{*}{\textbf{Setting}} 
           & \multicolumn{2}{c|}{\textbf{AIME 24}} 
           & \multicolumn{2}{c|}{\textbf{AIME 25}} 
           & \multicolumn{2}{c|}{\textbf{Beyond AIME}} 
           & \multicolumn{2}{c|}{\textbf{GPQA}} 
           & \multicolumn{2}{c|}{\textbf{MMLU}} 
           & \multicolumn{2}{c}{\textbf{Overall}} \\ \cmidrule(lr){2-13}
        & \texttt{avg@32} & \texttt{Toks}
        & \texttt{avg@32} & \texttt{Toks}
        & \texttt{avg@8} & \texttt{Toks}
        & \texttt{avg@8} & \texttt{Toks}
        & \texttt{avg@1} & \texttt{Toks}
        & \texttt{Avg.} & \texttt{Toks} \\
        \midrule
        \multicolumn{13}{c}{\texttt{\textbf{{DeepSeek-Distill-Qwen-1.5B}}}} \\
        \midrule
        Prompt Steering Low & 26.1 & 15.1 & 21.9 & 14.2 & 8.5 & 13.8 & 28.7 & 8.4 & 34.5 & 5.4 & 23.9 & 11.4 \\
        Prompt Steering Mid & 26.8 & 15.5 & 22.6 & 14.9 & 10.0 & 15.0 & 28.6 & 9.9 & 35.3 & 6.3 & 24.7 & 12.3 \\
        Prompt Steering High & 28.4 & 16.2 & 22.5 & 15.9 & 10.5 & 16.2 & 28.9 & 11.4 & 35.3 & 6.8 & 25.1 & 13.3 \\
        Prompt Steering Xhigh & 28.2 & 16.1 & 20.0 & 15.7 & 10.0 & 15.9 & 29.0 & 12.3 & 35.1 & 7.0 & 24.5 & 13.4 \\
        \midrule
        \multicolumn{13}{c}{\texttt{\textbf{{Qwen3-4B-2507-Thinking}}}} \\
        \midrule
        Prompt Steering Low & 79.3 & 16.6 & 77.0 & 17.1 & 47.1 & 20.4 & 63.6 & 7.3 & 73.1 & 3.8 & 68.0 & 13.0 \\
        Prompt Steering Mid & 79.0 & 19.9 & 77.3 & 20.4 & 54.4 & 26.8 & 66.7 & 8.5 & 72.8 & 4.6 & 70.0 & 16.0 \\
        Prompt Steering High & 81.1 & 20.4 & 77.7 & 20.8 & 53.5 & 27.7 & 65.4 & 9.1 & 74.1 & 5.0 & 70.4 & 16.6 \\
        Prompt Steering Xhigh & 81.9 & 21.1 & 79.1 & 21.9 & 53.6 & 28.8 & 66.9 & 9.9 & 73.7 & 6.2 & 71.0 & 17.6 \\
        \midrule
        \multicolumn{13}{c}{\texttt{\textbf{{Openmath-Nemotron-7B}}}} \\
        \midrule
        Prompt Steering Low & 70.8 & 11.3 & 57.5 & 13.7 & 41.3 & 13.1 & 31.9 & 8.2 & 42.6 & 5.5 & 48.8 & 10.4 \\
        Prompt Steering Mid & 70.0 & 11.8 & 57.5 & 13.1 & 40.3 & 13.6 & 33.4 & 8.2 & 42.4 & 5.7 & 48.7 & 10.5 \\
        Prompt Steering High & 73.3 & 11.1 & 55.8 & 13.3 & 41.8 & 13.5 & 34.4 & 8.6 & 42.5 & 5.7 & 49.6 & 10.4 \\
        Prompt Steering Xhigh & 69.1 & 12.1 & 61.0 & 13.1 & 42.6 & 13.3 & 34.1 & 8.5 & 42.3 & 5.8 & 49.8 & 10.6 \\
        \bottomrule[1.6pt]
    \end{tabular}}
    \caption{Prompt steering results on initial models across various reasoning tasks.}
    \label{tab:prompt_steering}
\end{table*}

\subsection{Frontier Teacher Model Performance}

We provide the evaluation results for each teacher model from exploration stage in Table~\ref{tab:teacher_overview}.

\subsection{Prompt Steering}
We evaluate prompt-based steering as a baseline to assess whether controllable behavior can be achieved through simple system prompts (e.g., instructing a vanilla LRM to be concise). This comparison quantifies the benefits of our specialized training (Table~\ref{tab:prompt_steering}).
Results show limited controllability: performance remains close to the default across effort levels, with no consistent mode-dependent scaling. Output length varies slightly, but accuracy changes are modest and often non-monotonic, indicating that ORBIT’s clear mode separation arises from specialized training rather than prompt steering.

\end{document}